\documentclass[10pt,twocolumn,letterpaper]{article}

\usepackage{cvpr}              %

\usepackage{graphicx}
\usepackage{amsmath}
\usepackage{amssymb}
\usepackage{booktabs}
\usepackage{multirow}
\usepackage[utf8]{inputenc} %
\usepackage[T1]{fontenc}    %
\usepackage[accsupp]{axessibility} 
\usepackage{float}

\usepackage{amsmath,amsfonts,bm}

\def\eqref#1{equation~\ref{#1}}

\def\1{\bm{1}}

\DeclareMathAlphabet{\mathsfit}{\encodingdefault}{\sfdefault}{m}{sl}
\SetMathAlphabet{\mathsfit}{bold}{\encodingdefault}{\sfdefault}{bx}{n}

\usepackage{color}
\usepackage[dvipsnames]{xcolor}
\usepackage{epsfig}

\usepackage{adjustbox}
\usepackage{array}
\usepackage{booktabs}
\usepackage{colortbl}
\usepackage{wrapfig}
\usepackage{hhline}
\usepackage{multirow}

\usepackage{paralist}
\usepackage{tabularx}

\usepackage{amsmath,amsfonts,amssymb}
\usepackage{bm}
\usepackage{nicefrac}
\usepackage{microtype}

\usepackage{changepage}
\usepackage{extramarks}
\usepackage{fancyhdr}
\usepackage{lastpage}
\usepackage{setspace}
\usepackage{soul}
\usepackage{xspace}

\usepackage{tabularx}

\usepackage{url}

\usepackage{enumerate}
\usepackage{enumitem}  %

\usepackage{makecell}

\usepackage{pifont} %
\usepackage[ruled,vlined]{algorithm2e}
\usepackage{tabularx}

\newcolumntype{L}[1]{>{\raggedright\let\newline\\\arraybackslash\hspace{0pt}}m{#1}}
\newcolumntype{C}[1]{>{\centering\let\newline\\\arraybackslash\hspace{0pt}}m{#1}}
\newcolumntype{R}[1]{>{\raggedleft\let\newline\\\arraybackslash\hspace{0pt}}m{#1}}

\newcommand{\ignore}[1]{}

\makeatletter
\DeclareRobustCommand\onedot{\futurelet\@let@token\@onedot}
\def\@onedot{\ifx\@let@token.\else.\null\fi\xspace}

\def\etal{et al\onedot}

\makeatother

\definecolor{MyDarkBlue}{rgb}{0,0.08,1}
\definecolor{MyDarkGreen}{rgb}{0.02,0.6,0.02}
\definecolor{MyDarkRed}{rgb}{0.8,0.02,0.02}
\definecolor{MyDarkOrange}{rgb}{0.40,0.2,0.02}
\definecolor{MyPurple}{RGB}{111,0,255}
\definecolor{MyRed}{rgb}{1.0,0.0,0.0}
\definecolor{MyGold}{rgb}{0.75,0.6,0.12}
\definecolor{MyDarkgray}{rgb}{0.66, 0.66, 0.66}

\definecolor{bittersweet}{rgb}{1.0, 0.44, 0.37}
\definecolor{ballblue}{rgb}{0.13, 0.67, 0.8}
\definecolor{amethyst}{rgb}{0.6, 0.4, 0.8}
\definecolor{blue-violet}{rgb}{0.54, 0.17, 0.89}
\definecolor{brightpink}{rgb}{1.0, 0.0, 0.5}

\newcommand{\xhdr}[1]{\noindent\textbf{#1}}
\usepackage{subcaption}

\usepackage[pagebackref,breaklinks,colorlinks]{hyperref}

\usepackage[capitalize]{cleveref}
\crefname{section}{Sec.}{Secs.}
\Crefname{section}{Section}{Sections}
\Crefname{table}{Table}{Tables}
\crefname{table}{Tab.}{Tabs.}

\def\cvprPaperID{737} %
\def\confName{CVPR}
\def\confYear{2023}

\begin{document}

\title{Putting People in Their Place: Affordance-Aware Human Insertion into Scenes\vspace{-0.5em}}
\author{
Sumith Kulal$^{1}$\\
\vspace{-1em}
\and
Tim Brooks$^{2}$
\and
Alex Aiken$^{1}$
\and
Jiajun Wu$^{1}$
\and
Jimei Yang$^{3}$
\and
Jingwan Lu$^{3}$
\and
Alexei A.\ Efros$^{2}$
\and
Krishna Kumar Singh$^{3}$
}

\twocolumn[{
\maketitle
\vspace{-2.8em}
\begin{center}
{\large
$^{1}$Stanford University\hspace{0.8em}$^{2}$UC Berkeley\hspace{0.8em}$^{3}$Adobe Research
}
\end{center}
\vspace{1em}
}]
\begin{figure*}
\centering
    \centering

    \centering
        \begin{minipage}{\linewidth}
        \includegraphics[width=.124\linewidth]{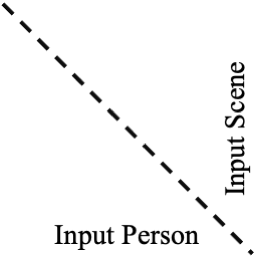}
        \hspace{-0.1em}
        \includegraphics[width=.124\linewidth]{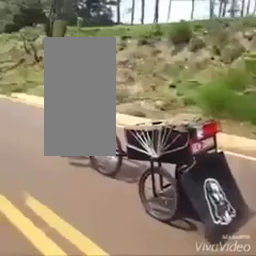}
        \hspace{-0.4em}
        \includegraphics[width=.124\linewidth]{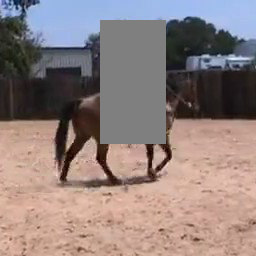}
        \hspace{-0.4em}
        \includegraphics[width=.124\linewidth]{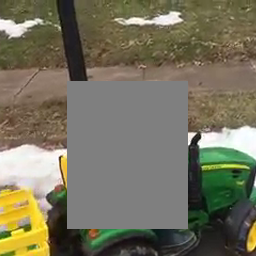}
        \hspace{-0.4em}
        \includegraphics[width=.124\linewidth]{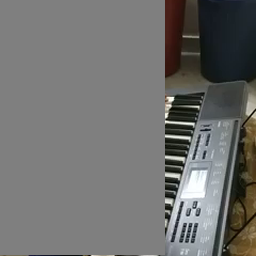}
        \hspace{-0.4em}
        \includegraphics[width=.124\linewidth]{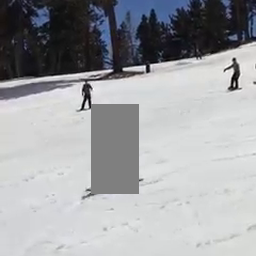}
        \hspace{-0.4em}
        \includegraphics[width=.124\linewidth]{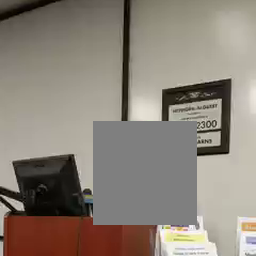}
        \hspace{-0.4em}
        \includegraphics[width=.124\linewidth]{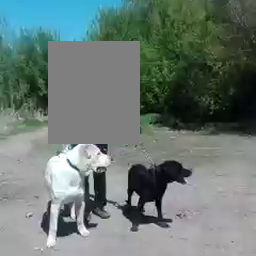}
        \end{minipage}

\vspace{0.15em}

    \centering
        \begin{minipage}{\linewidth}
        \includegraphics[width=.124\linewidth]{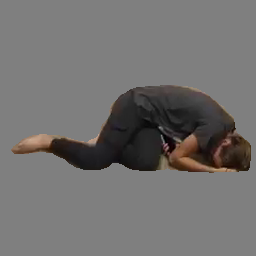}
        \hspace{-0.2em}
        \includegraphics[width=.124\linewidth]{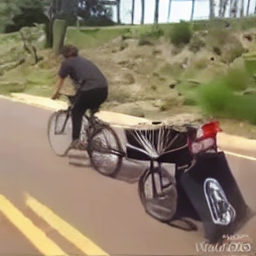}
        \hspace{-0.4em}
        \includegraphics[width=.124\linewidth]{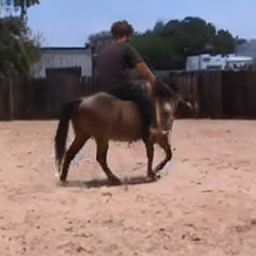}
        \hspace{-0.4em}
        \includegraphics[width=.124\linewidth]{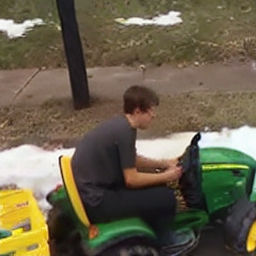}
        \hspace{-0.4em}
        \includegraphics[width=.124\linewidth]{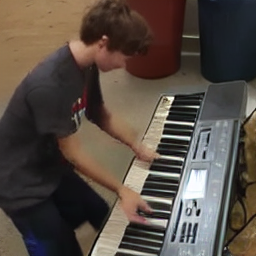}
        \hspace{-0.4em}
        \includegraphics[width=.124\linewidth]{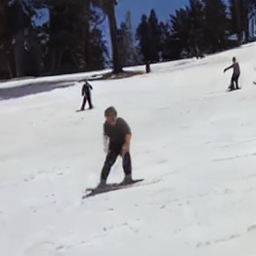}
        \hspace{-0.4em}
        \includegraphics[width=.124\linewidth]{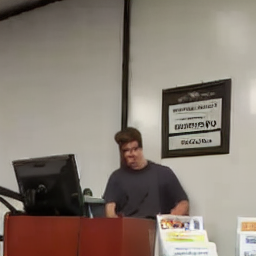}
        \hspace{-0.4em}
        \includegraphics[width=.124\linewidth]{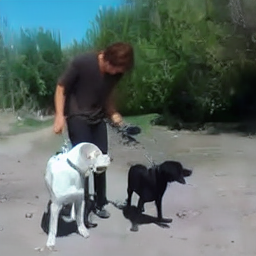}
        \end{minipage}
    \centering
        \begin{minipage}{\linewidth}
        \includegraphics[width=.124\linewidth]{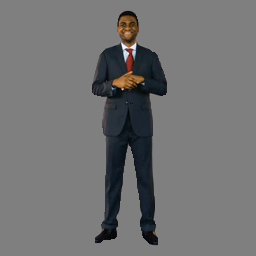}
        \hspace{-0.2em}
        \includegraphics[width=.124\linewidth]{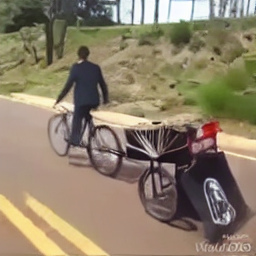}
        \hspace{-0.4em}
        \includegraphics[width=.124\linewidth]{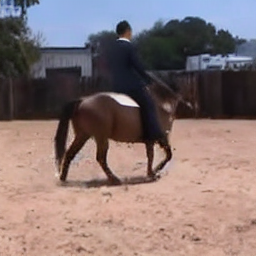}
        \hspace{-0.4em}
        \includegraphics[width=.124\linewidth]{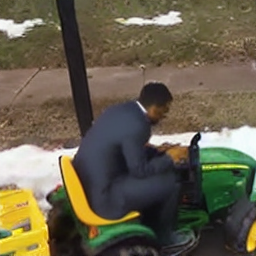}
        \hspace{-0.4em}
        \includegraphics[width=.124\linewidth]{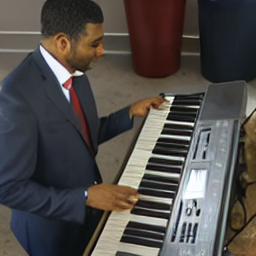}
        \hspace{-0.4em}
        \includegraphics[width=.124\linewidth]{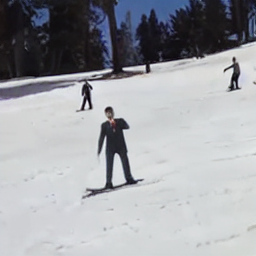}
        \hspace{-0.4em}
        \includegraphics[width=.124\linewidth]{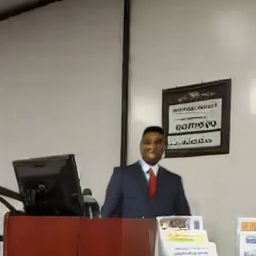}
        \hspace{-0.4em}
        \includegraphics[width=.124\linewidth]{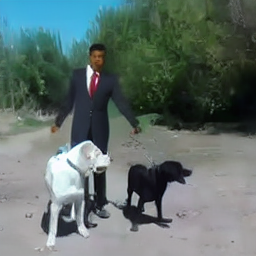}
        \end{minipage}
    \centering
        \begin{minipage}{\linewidth}
        \includegraphics[width=.124\linewidth]{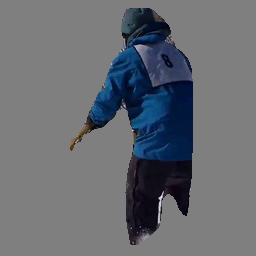}
        \hspace{-0.2em}
        \includegraphics[width=.124\linewidth]{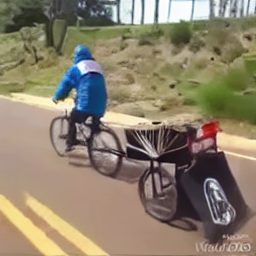}
        \hspace{-0.4em}
        \includegraphics[width=.124\linewidth]{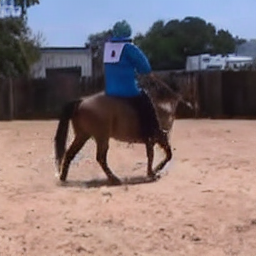}
        \hspace{-0.4em}
        \includegraphics[width=.124\linewidth]{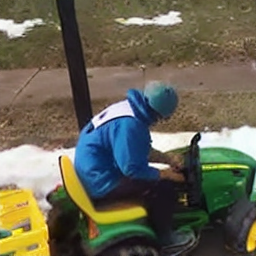}
        \hspace{-0.4em}
        \includegraphics[width=.124\linewidth]{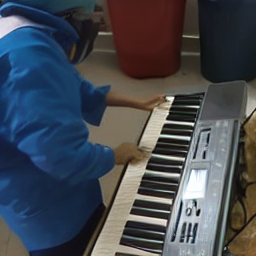}
        \hspace{-0.4em}
        \includegraphics[width=.124\linewidth]{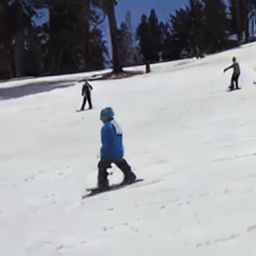}
        \hspace{-0.4em}
        \includegraphics[width=.124\linewidth]{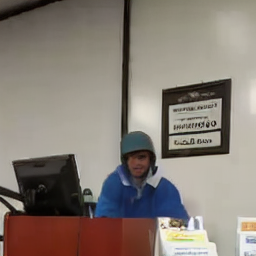}
        \hspace{-0.4em}
        \includegraphics[width=.124\linewidth]{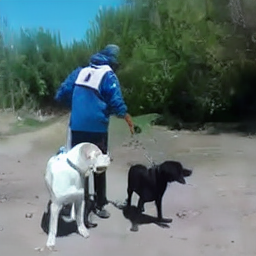}
        \end{minipage}
    \centering
        \begin{minipage}{\linewidth}
        \includegraphics[width=.124\linewidth]{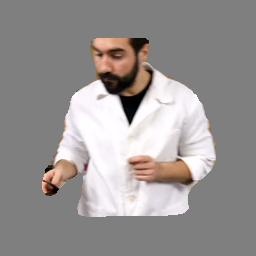}
        \hspace{-0.2em}
        \includegraphics[width=.124\linewidth]{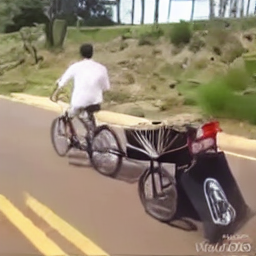}
        \hspace{-0.4em}
        \includegraphics[width=.124\linewidth]{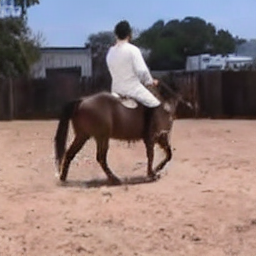}
        \hspace{-0.4em}
        \includegraphics[width=.124\linewidth]{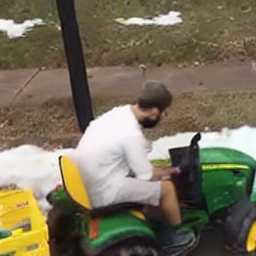}
        \hspace{-0.4em}
        \includegraphics[width=.124\linewidth]{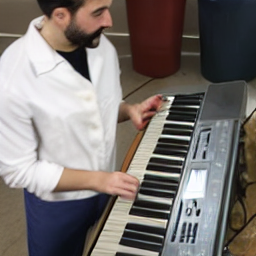}
        \hspace{-0.4em}
        \includegraphics[width=.124\linewidth]{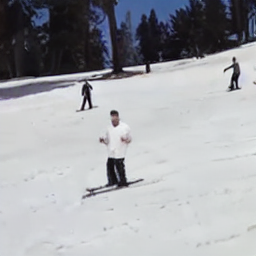}
        \hspace{-0.4em}
        \includegraphics[width=.124\linewidth]{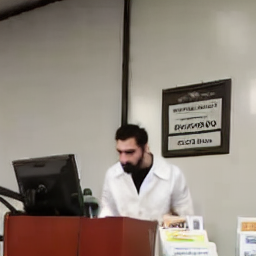}
        \hspace{-0.4em}
        \includegraphics[width=.124\linewidth]{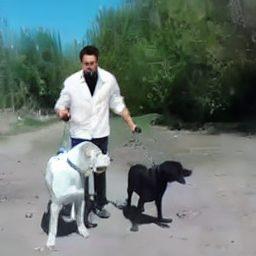}
        \end{minipage}
    \centering
        \begin{minipage}{\linewidth}
        \includegraphics[width=.124\linewidth]{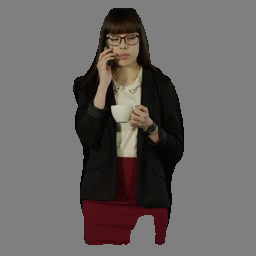}
        \hspace{-0.2em}
        \includegraphics[width=.124\linewidth]{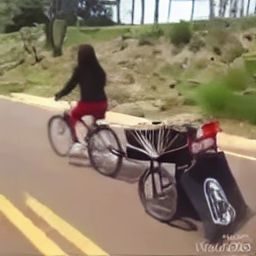}
        \hspace{-0.4em}
        \includegraphics[width=.124\linewidth]{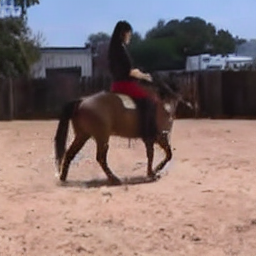}
        \hspace{-0.4em}
        \includegraphics[width=.124\linewidth]{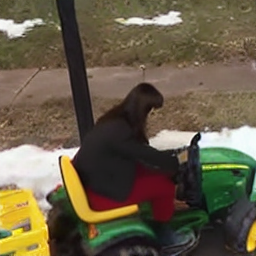}
        \hspace{-0.4em}
        \includegraphics[width=.124\linewidth]{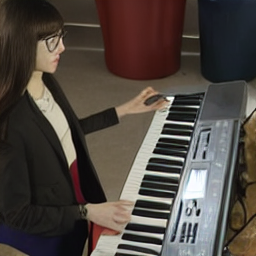}
        \hspace{-0.4em}
        \includegraphics[width=.124\linewidth]{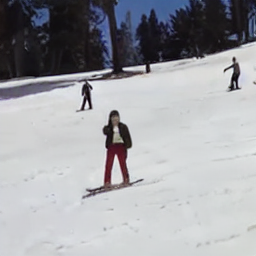}
        \hspace{-0.4em}
        \includegraphics[width=.124\linewidth]{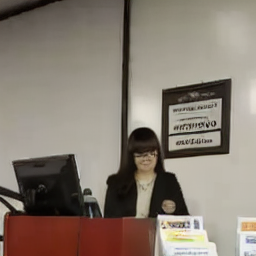}
        \hspace{-0.4em}
        \includegraphics[width=.124\linewidth]{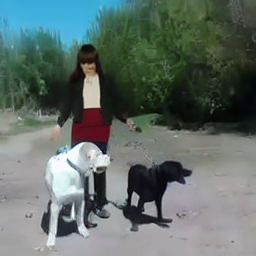}
        \end{minipage}
    \centering
        \begin{minipage}{\linewidth}
        \includegraphics[width=.124\linewidth]{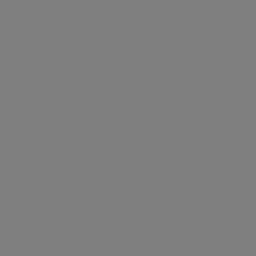}
        \hspace{-0.2em}
        \includegraphics[width=.124\linewidth]{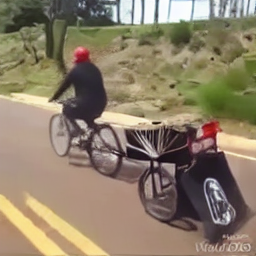}
        \hspace{-0.4em}
        \includegraphics[width=.124\linewidth]{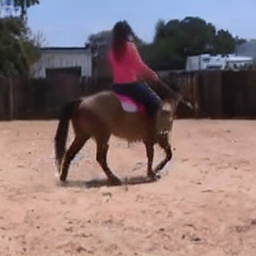}
        \hspace{-0.4em}
        \includegraphics[width=.124\linewidth]{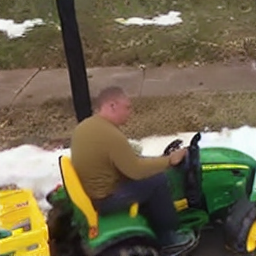}
        \hspace{-0.4em}
        \includegraphics[width=.124\linewidth]{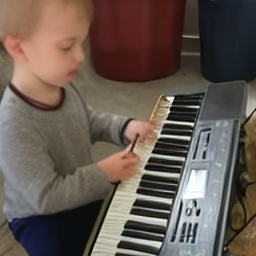}
        \hspace{-0.4em}
        \includegraphics[width=.124\linewidth]{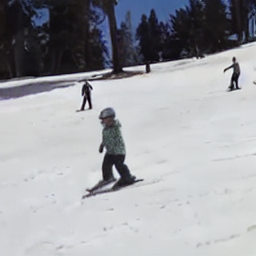}
        \hspace{-0.4em}
        \includegraphics[width=.124\linewidth]{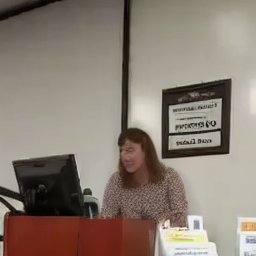}
        \hspace{-0.4em}
        \includegraphics[width=.124\linewidth]{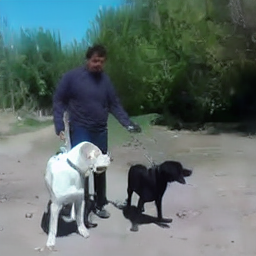}
        \end{minipage}

\vspace{-0.4em}
\caption{
Given a masked scene image (first row) and a reference person (first column), our model can successfully insert the person into the scene image. The model infers the possible pose (affordance) given the scene context, reposes the person appropriately, and harmonizes the insertion. We can also partially complete a person (last column) and hallucinate a person (last row) when no reference is given. }
\vspace{-0.4em}

\label{fig:teaser}
\end{figure*}

\begin{abstract}
\vspace{-0.2em}

We study the problem of inferring scene affordances by presenting a method for realistically inserting people into scenes. Given a scene image with a marked region and an image of a person, we insert the person into the scene while respecting the scene affordances. 
Our model can infer the set of realistic poses given the scene context, re-pose the reference person, and harmonize the composition.  We set up the task in a self-supervised fashion by learning to re-pose humans in video clips. 
We train a large-scale diffusion model on a dataset of 2.4M video clips that produces diverse plausible poses while respecting the scene context. Given the learned human-scene composition, our model can also hallucinate realistic people and scenes when prompted without conditioning and also enables interactive editing. 
A quantitative evaluation shows that our method synthesizes more realistic human appearance and more natural human-scene interactions than prior work. 

\end{abstract}

\vspace{-1.5em}
\section{Introduction}
\vspace{-0.4em}
\label{sec:intro}

A hundred years ago, Jakob von Uexküll 
pointed out the crucial, even defining, role that the perceived environment ({\em umwelt}) plays in an organism's life~\cite{uexkull1985environment}.  At a high level, he argued that an organism is only aware of the parts of the environment that it can affect or be affected by. In a sense, our perception of the world is defined by what kinds of interactions we can perform.
Related ideas of functional visual understanding (what actions does a given scene afford an agent?) 
were discussed in the 1930s by the Gestalt psychologists~\cite{koffka1935principles} and later described by J.J.\ Gibson~\cite{gibson1979} as {\em affordances}. Although this direction inspired many efforts in vision and psychology research, a comprehensive computational model of affordance perception remains elusive. The value of such a computational model is undeniable for future work in vision and robotics research.
\footnotetext{Project page: \url{https://sumith1896.github.io/affordance-insertion}.}

The past decade has seen a renewed interest in such computational models for data-driven affordance perception~\cite{GuptaSatkinEfrosHebertCVPR11, delaitre2012, fouhey2015defense, WangaffordanceCVPR2017, Hassan2021}. Early works in this space deployed a mediated approach by inferring or using intermediate semantic or 3D information to aid in affordance perception~\cite{GuptaSatkinEfrosHebertCVPR11}, while more recent methods focus on direct perception of affordances~\cite{delaitre2012,fouhey2015defense, WangaffordanceCVPR2017}, more in line with Gibson's framing~\cite{gibson1979}. However, these methods are severely constrained by the specific requirements of the datasets, which reduce their generalizability.

To facilitate a more general setting, we draw inspiration from the recent advances in large-scale generative models, such as text-to-image systems~\cite{saharia2022photorealistic, ramesh2022hierarchical, rombach2022high}. The samples from these models demonstrate impressive object-scene compositionality. However, these compositions are implicit, and the affordances are limited to what is typically captured in still images and described by captions. We make the task of affordance prediction explicit by putting people ``into the picture''~\cite{GuptaSatkinEfrosHebertCVPR11} and training on videos of human activities.

We pose our problem as a conditional inpainting task (Fig.~\ref{fig:teaser}). Given a masked scene image (first row) and a reference person (first column), we learn to inpaint the person into the masked region with correct affordances. At training time, we borrow two random frames from a video clip, mask one frame, and try to inpaint using the person from the second frame as the condition. This forces the model to learn both the possible scene affordances given the context and the necessary re-posing and harmonization needed for a coherent image. At inference time, the model can be prompted with different combinations of scene and person images. We train a large-scale model on a dataset of 2.4M video clips of humans moving in a wide variety of scenes.

In addition to the conditional task, our model can be prompted in different ways at inference time. As shown in the last row Fig.~\ref{fig:teaser}, when prompted without a person, our model can hallucinate a realistic person. Similarly, when prompted without a scene, it can also hallucinate a realistic scene. One can also perform partial human completion tasks such as changing the pose or swapping clothes. We show that training on videos is crucial for predicting affordances and present ablations and baseline comparisons in Sec.~\ref{sec:experiments}.

To summarize, our contributions are: 
\begin{itemize}[topsep=0pt,parsep=0pt,partopsep=0pt,leftmargin=2em]
\setlength\itemsep{0.1em}
\item We present a fully self-supervised task formulation for learning affordances by learning to inpaint humans in masked scenes.
\item We present a large-scale generative model for human insertion trained on 2.4M video clips and demonstrate improved performance both qualitatively and quantitatively compared to the baselines.
\item In addition to conditional generation, our model can be prompted in multiple ways to support person hallucination, scene hallucination, and interactive editing.  
\end{itemize}

\vspace{-0.2em}
\section{Related Work}
\vspace{-0.2em}
\label{sec:related-works}
\xhdr{Scene and object affordances.} Inspired by the work of J.J. Gibson~\cite{gibson1979}, a long line of papers have looked into operationalizing affordance prediction~\cite{GuptaSatkinEfrosHebertCVPR11,Grabner2011chair,fouhey2012people,delaitre2012,Jiang2013CVPR,fouhey2015defense,WangaffordanceCVPR2017,chuang2018learning,3d-affordance, brooks2021hallucinating}. Prior works have also looked at modeling human-object affordance~\cite{Yao2010ModelingMC,koppula2013learning,zhu2014reasoning,gkioxari2018detecting,cao2020reconstructing} and synthesizing human pose (and motion) conditioned on an input scene~\cite{lee2002interactive,caoHMP2020,wang2020synthesizing}. Several methods have used videos of humans interacting with scenes to learn about scene affordances~\cite{fouhey2012people,delaitre2012,WangaffordanceCVPR2017}. For example, Wang~\etal~\cite{WangaffordanceCVPR2017} employed a large-scale video dataset to directly predict affordances. They generated a dataset of possible human poses in sitcom scenes. However, their model relies on having plausible ground-truth poses for scenes and hence only performs well on a small number of scenes and poses. On the other hand, we work with a much larger dataset and learn affordances in a fully self-supervised generative manner. We also go beyond synthesizing pose alone and generate realistic humans conditioned on the scene. By virtue of scale, our work generalizes better to diverse scenes and poses and could be scaled further~\cite{sutton2019bitter}. 

\xhdr{Inpainting and hole-filling.} Early works attempted to use the information within a single image to inpaint masked regions by either diffusing local appearance~\cite{bertalmio2000image, osher2005iterative, ballester2001filling} or matching patches~\cite{efros1999texture, barnes2009patchmatch}. More recent works use larger datasets to match features~\cite{hays2007scene, pathak2016context}. Pathak~\etal~\cite{pathak2016context} showed a learning-based approach for image inpainting for large masks, followed up by several recent works that use CNNs~\cite{yang2017high, iizuka2017globally, yu2018generative, liu2018image, yu2019free, zhao2021large, zheng2022cm} and Transformers~\cite{esser2021taming, yu2021diverse, bar2022visual}. The most relevant works to ours are diffusion-based inpainting models~\cite{rombach2022high, saharia2022palette, lugmayr2022repaint}. Rombach~\etal~\cite{rombach2022high} used text to guide the diffusion models to perform inpainting tasks. Our task can also be considered as guided inpainting, but our conditioning is an image of a person to be inserted in the scene instead of text. The masking strategy we use is  inspired by~\cite{zhao2021large, yu2019free}.

\xhdr{Conditional human synthesis.} Several works have attempted synthesizing human images (and videos) from conditional information such as keypoints~\cite{balakrishnan2018synthesizing, Li2019CVPR, ma2017pose, siarohin2018deformable, aberman2019deep, chan2019everybody, villegas2017learning}, segmentation masks or densepose~\cite{neverova2018dense, wang2018vid2vid, albahar2021pose, yoon2021pose}, and driving videos~\cite{tulyakov2018mocogan, siarohin2019first}. Prior reposing works do not consider scene context to infer the pose, since the target pose is explicitly given. Moreover, most of the reposing happens in simple backgrounds without semantic content. In contrast, our model conditions on the input scene context and infers the right pose (affordance) prior to reposing. Additionally, our model is trained on unconstrained real-world scenes in an end-to-end manner with no explicit intermediate representation, such as keypoints or 3D.

\xhdr{Diffusion models.} Introduced as an expressive and powerful generative model~\cite{sohl2015deep}, diffusion models have been shown to outperform GANs~\cite{ho2020denoising, nichol2021improved, dhariwal2021diffusion} in generating more photorealistic and diverse images unconditionally or conditioned by text. 
With a straightforward architecture, they achieve promising performance in several text-to-image~\cite {nichol2021glide, ramesh2022hierarchical, saharia2022photorealistic, rombach2022high}, video~\cite{ho2022imagen, singer2022make}, and 3D synthesis~\cite{poole2022dreamfusion} tasks. 
We leverage ideas presented by Rombach~\etal~\cite{rombach2022high} which first encodes images into a latent space and then performs diffusion training in the latent space. 
We also use classifier-free guidance, introduced by Ho~\etal~\cite{ho2022classifier}, which 
improves sample quality by trading off against diversity.
\vspace{-0.2em}
\section{Methods}
\label{sec:framework}
\vspace{-0.2em}
In this section, we present details of our learning framework. Given an input scene image, a masked region, and a reference person to be inserted, our model inpaints the masked region with a photo-realistic human that follows the appearance of the reference person, but is  re-posed to fit the context in the input scene. We use the latent diffusion model as our base architecture, described in Sec.~\ref{sec:diffusion-models}. We present details on our problem formulation in Sec.~\ref{sec:formulation}, our training data in Sec.~\ref{sec:training-data}, and our model in Sec.~\ref{sec:model-details}.

\begin{figure*}
\centering
\includegraphics[width=\linewidth]{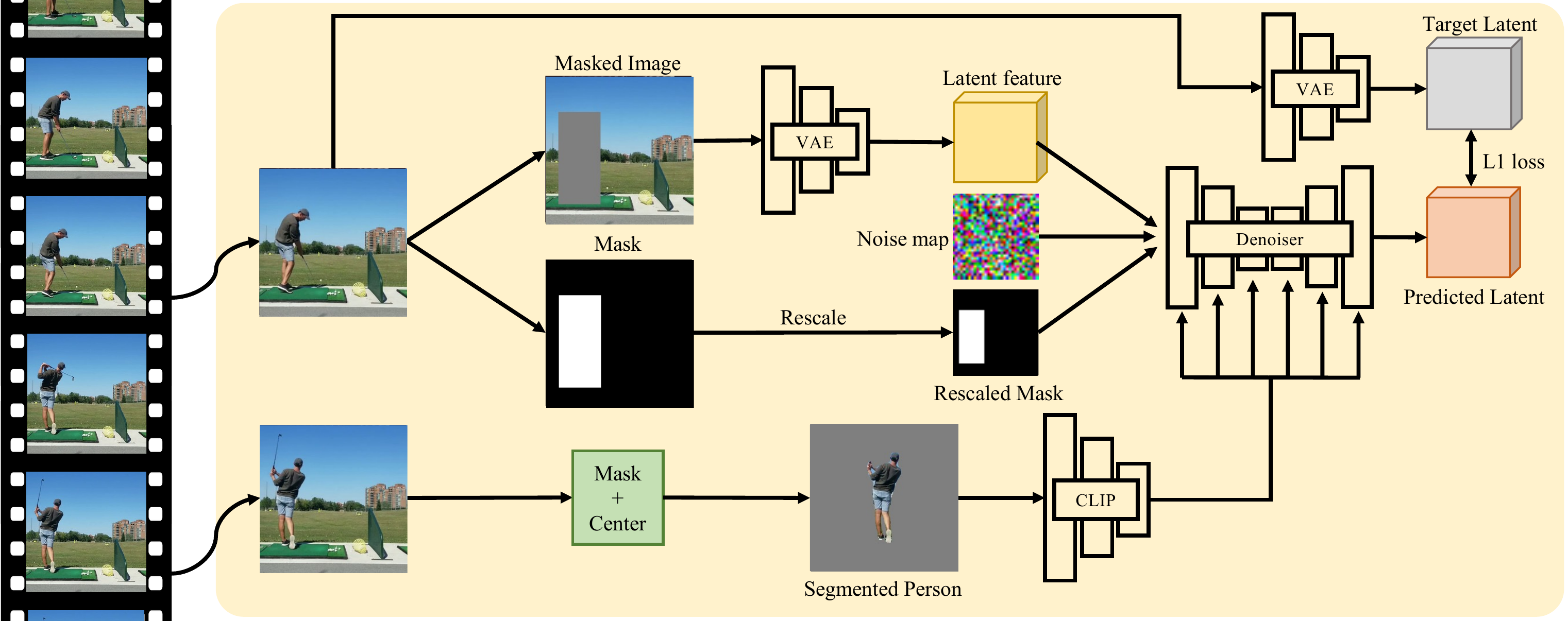}
\vspace{-0.5mm}
\caption{
\textbf{Architecture overview.} We source two random frames from a video clip. We mask out the person in the first frame and use the person from the second frame as conditioning to inpaint the image. We concatenate the latent features of the background image and rescaled mask along with the noisy image to the denoising UNet. Reference person embeddings (CLIP ViT-L/14) are passed via cross-attention. }
\label{fig:arch}
\end{figure*}

\vspace{-0.2em}
\subsection{Background - Diffusion Models}
\label{sec:diffusion-models}
\vspace{-0.2em}
Diffusion models~\cite{sohl2015deep, ho2020denoising} are generative models that model data distribution $p(x)$ as a sequence of denoising autoencoders. For a fixed time step $T$, the forward process of diffusion models gradually adds noise in $T$ steps to destroy the data signal. At time $T$ the samples are approximately uniform Gaussian noise. The reverse process then learns to denoise into samples in $T$ steps. These models effectively predict $\epsilon_{\theta}(x_t,t)$ for $t=1 \dots T$, the noise-level at time-step $t$ given the $x_t$, a noisy version of input $x$. The corresponding simplified training objective~\cite{rombach2022high} is
\begin{equation}
    L_{DM} = \mathbb{E}_{x,\epsilon\sim\mathcal{N}(0,1),t} \Bigl[\|\epsilon -  \epsilon_{\theta}(x_t,t,c) \|_2^2  \Bigr],
\label{eq:dm}
\end{equation}
where $t$ is uniformly sampled from ${\{1,\dots,T}\}$ and $c$ are the conditioning variables: the masked scene image and the reference person.

\xhdr{Latent diffusion models.} As shown in Rombach \etal~\cite{rombach2022high}, we use an autoencoder to do perceptual compression and let the diffusion model focus on the semantic content, which makes the training more computationally efficient. Given an autoencoder with encoder $\mathcal{E}$ and decoder $\mathcal{D}$, the forward process uses $\mathcal{E}$ to encode the image, and samples from the model are decoded using $\mathcal{D}$ back to the pixel space.

\xhdr{Classifier-free guidance.} Ho \etal~\cite{ho2022classifier} proposed classifier-free guidance (CFG) for trading off sample quality with diversity. The idea is to amplify the difference between conditional and unconditional prediction during sampling for the same noisy image. The updated noise prediction is
\begin{equation}
    \hat{\epsilon} = w\cdot\epsilon_{\theta}(x_t,t,c) - (w - 1)\cdot\epsilon_{\theta}(x_t,t),
\label{eq:cfg}
\end{equation}

\vspace{-0.2em}
\subsection{Formulation}
\vspace{-0.2em}
\label{sec:formulation}

The inputs to our model contain a masked scene image and a reference person, and the output image contains the reference person re-posed in the scene's context. 

Inspired by Humans in Context (HiC)~\cite{brooks2021hallucinating}, we generate a large dataset of videos with humans moving in scenes and use frames of videos as training data in our fully self-supervised training setup. We pose the problem as a conditional generation problem (shown in Fig.~\ref{fig:arch}). At training time, we source two random frames containing the same human from a video. We mask out the person in the first frame and use it as the input scene. We then crop out and center the human from the second frame and use it as the reference person conditioning. We train a conditional latent diffusion model conditioned on both the masked scene image and the reference person image. This encourages the model to infer the right pose given the scene context, hallucinate the person-scene interactions, and harmonize the re-posed person into the scene seamlessly in a self-supervised manner. 

At test time, the model can support multiple applications, inserting different reference humans, hallucinating humans without references, and hallucinating scenes given the human. We achieve this by randomly dropping conditioning signals during training. We evaluate the quality of person conditioned generation, person hallucination and scene hallucination in our experimental section.

\vspace{-0.2em}
\subsection{Training data}
\begin{figure}
    \centering
    \begin{subfigure}{\linewidth}
    \centering
        \begin{minipage}{\linewidth}
            \includegraphics[width=.195\linewidth]{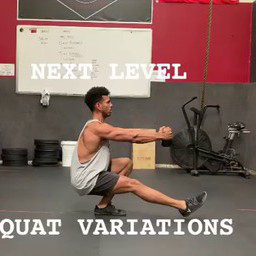}
            \hspace{-0.4em}
            \includegraphics[width=.195\linewidth]{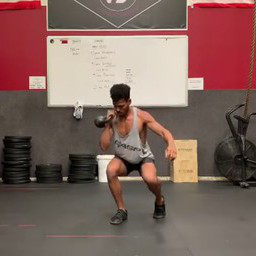}
            \hspace{-0.4em}
            \includegraphics[width=.195\linewidth]{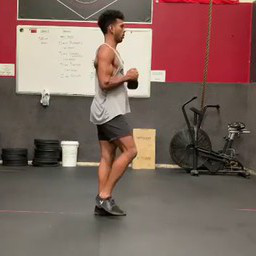}
            \hspace{-0.4em}
            \includegraphics[width=.195\linewidth]{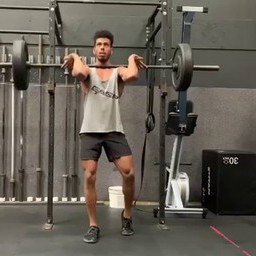}
            \hspace{-0.4em}
            \includegraphics[width=.195\linewidth]{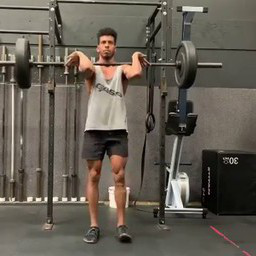}
        \end{minipage}
    \end{subfigure}
    \begin{subfigure}{\linewidth}
    \centering
        \begin{minipage}{\linewidth}
            \includegraphics[width=.195\linewidth]{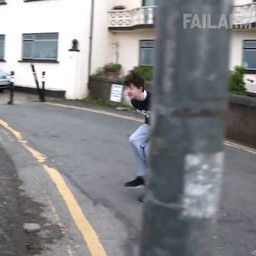}
            \hspace{-0.4em}
            \includegraphics[width=.195\linewidth]{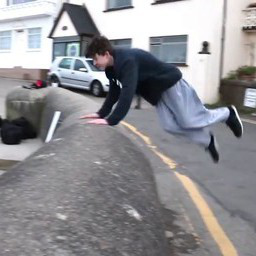}
            \hspace{-0.4em}
            \includegraphics[width=.195\linewidth]{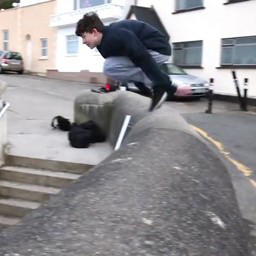}
            \hspace{-0.4em}
            \includegraphics[width=.195\linewidth]{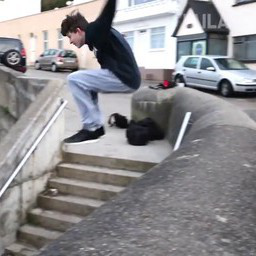}
            \hspace{-0.4em}
            \includegraphics[width=.195\linewidth]{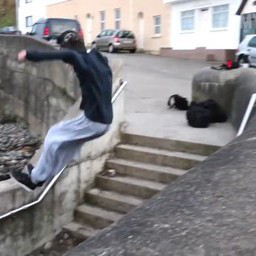}
        \end{minipage}
    \end{subfigure}
    \begin{subfigure}{\linewidth}
    \centering
        \begin{minipage}{\linewidth}
            \includegraphics[width=.195\linewidth]{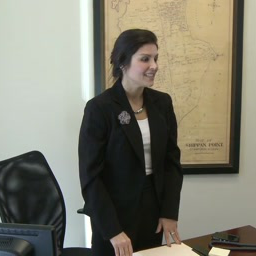}
            \hspace{-0.4em}
            \includegraphics[width=.195\linewidth]{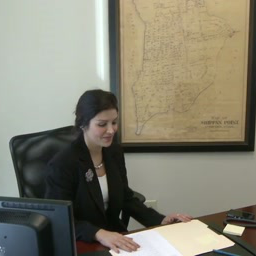}
            \hspace{-0.4em}
            \includegraphics[width=.195\linewidth]{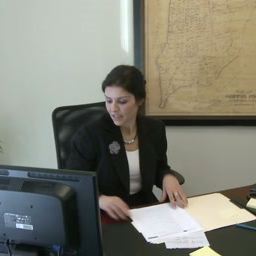}
            \hspace{-0.4em}
            \includegraphics[width=.195\linewidth]{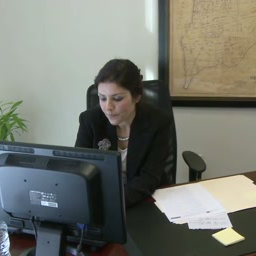}
            \hspace{-0.4em}
            \includegraphics[width=.195\linewidth]{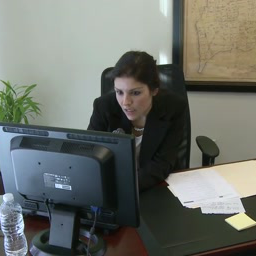}
        \end{minipage}
    \end{subfigure}
\vspace{-0.5em}
\caption{\textbf{
Sample videos from our dataset.} Each row has five frames uniformly sampled from a video.
\vspace{-0.5em}}
\label{fig:data}
\end{figure}

\vspace{-0.2em}
\label{sec:training-data}
We generate a dataset of 2.4 million video clips of humans moving in scenes. We follow the pre-processing pipeline defined in HiC~\cite{brooks2021hallucinating}. We start from around 12M videos, including a combination of publicly available computer vision datasets as in Brooks~\etal\cite{brooks2021hallucinating} and proprietary datasets. First, we resize all videos to a shorter-edge resolution of $256$ pixels and retain $256\times256$ cropped segments with a single person detected by Keypoint R-CNN~\cite{he2017mask}. We then filter out videos where OpenPose~\cite{cao2019openpose} does not detect a sufficient number of keypoints. This results in 2.4M videos, of which 50,000 videos are held out as the validation set, and the rest are used for training. Samples from the dataset are shown in Fig.~\ref{fig:data}. Finally, we use Mask R-CNN~\cite{he2017mask} to detect person masks to mask out humans in the input scene image and to crop out humans to create the reference person.

\begin{figure}
\vspace{-0.5em}
\centering

    \begin{subfigure}{\linewidth}
    \centering
        \begin{minipage}{\linewidth}
            \includegraphics[width=.166\linewidth]{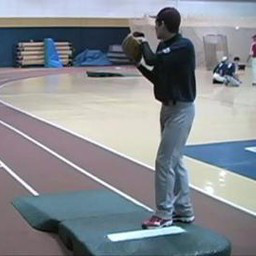}
            \hspace{-0.4em}
            \includegraphics[width=.166\linewidth]{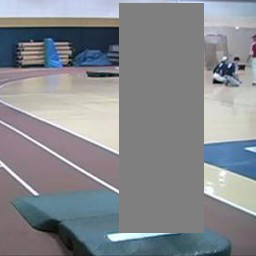}
            \hspace{-0.4em}
            \includegraphics[width=.166\linewidth]{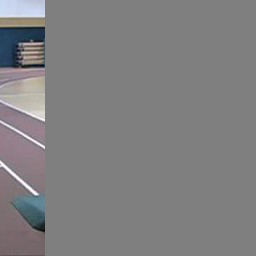}
            \hspace{-0.4em}
            \includegraphics[width=.166\linewidth]{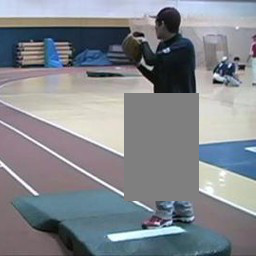}
            \hspace{-0.4em}
            \includegraphics[width=.166\linewidth]{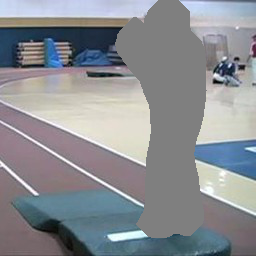}
            \hspace{-0.4em}
            \includegraphics[width=.166\linewidth]{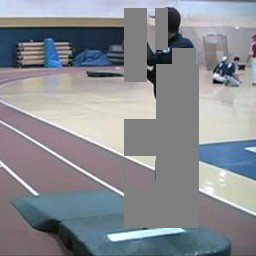}
        \end{minipage}
        \vspace{-0.18em}
    \end{subfigure}
    \begin{subfigure}{\linewidth}
    \centering
        \begin{minipage}{\linewidth}
            \includegraphics[width=.166\linewidth]{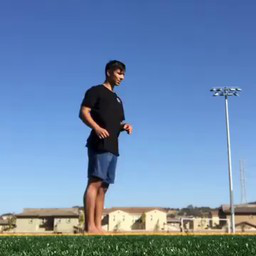}
            \hspace{-0.4em}
            \includegraphics[width=.166\linewidth]{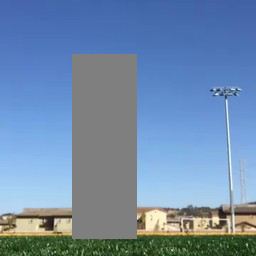}
            \hspace{-0.4em}
            \includegraphics[width=.166\linewidth]{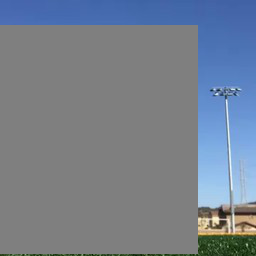}
            \hspace{-0.4em}
            \includegraphics[width=.166\linewidth]{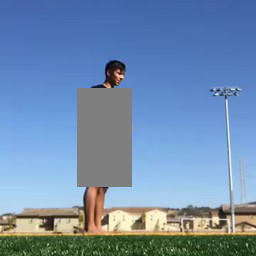}
            \hspace{-0.4em}
            \includegraphics[width=.166\linewidth]{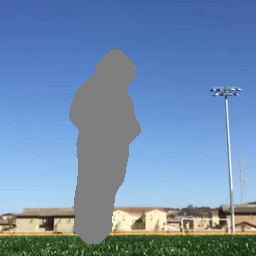}
            \hspace{-0.4em}
            \includegraphics[width=.166\linewidth]{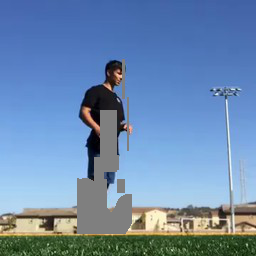}
        \end{minipage}
        \vspace{-0.18em}
    \end{subfigure}

   \vspace{-0.2em}
\caption{\textbf{Various masks used during training.} We use bounding boxes of the person, larger boxes around the person, smaller boxes, segmentation masks, and randomly generated scribbles. 
}
  \vspace{-1.4em}
\label{fig:mask}
\end{figure}

We briefly describe our masking and augmentation strategy and present more details in the supp.\ materials. 

\xhdr{Masking strategy.} We apply a combination of different masks for the input scene image,  as shown in Fig.~\ref{fig:mask}. These contain bounding boxes, segmentation masks and random scribbles as done in prior inpainting works~\cite{zhao2021large, yu2019free}. This masking strategy allows us to insert people at different levels of granularity, i.e., inserting the whole person, partially completing a person, etc.

\begin{figure}
\centering
    \begin{subfigure}{\linewidth}
    \centering
        \begin{minipage}{\linewidth}
            \includegraphics[width=.195\linewidth]{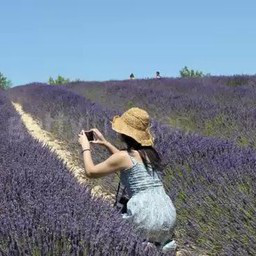}
            \hspace{-0.4em}
            \includegraphics[width=.195\linewidth]{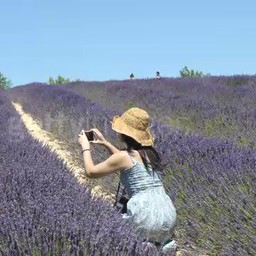}
            \hspace{-0.4em}
            \includegraphics[width=.195\linewidth]{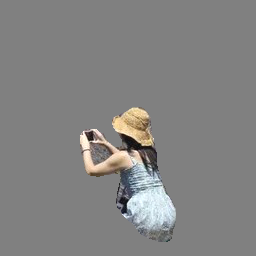}
            \hspace{-0.4em}
            \includegraphics[width=.195\linewidth]{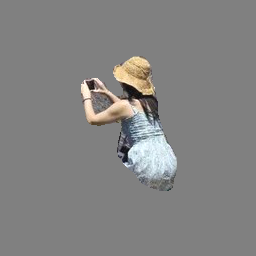}
            \hspace{-0.4em}
            \includegraphics[width=.195\linewidth]{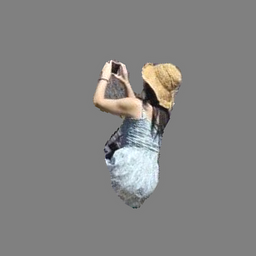}
        \end{minipage}
    \end{subfigure}
    \begin{subfigure}{\linewidth}
    \centering
        \begin{minipage}{\linewidth}
            \includegraphics[width=.195\linewidth]{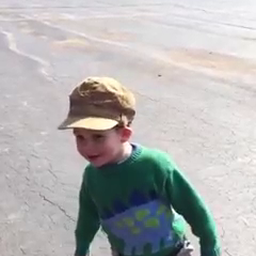}
            \hspace{-0.4em}
            \includegraphics[width=.195\linewidth]{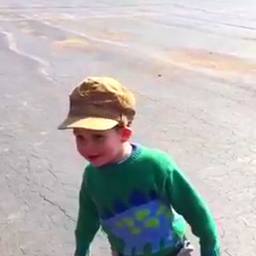}
            \hspace{-0.4em}
            \includegraphics[width=.195\linewidth]{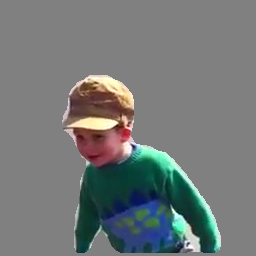}
            \hspace{-0.4em}
            \includegraphics[width=.195\linewidth]{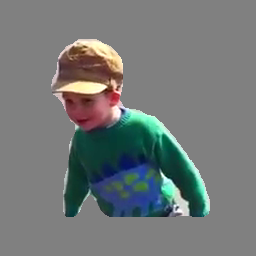}
            \hspace{-0.4em}
            \includegraphics[width=.195\linewidth]{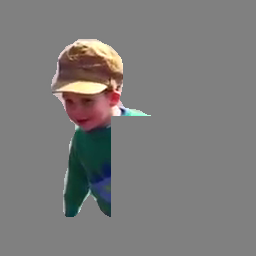}
        \end{minipage}
    \end{subfigure}
\caption{\textbf{Data augmentation used for the reference person.} We first apply color augmentations and image corruptions. We then mask and center the person, followed by geometric augmentations.
}
  \vspace{-1.4em}
\label{fig:aug}
\end{figure}
\xhdr{Augmentation strategy.} We apply data augmentation to reference person alone (as shown in Fig.~\ref{fig:aug}). We borrow the augmentation suite used in StyleGAN-ADA~\cite{karras2020training}. We randomly apply color augmentations. We then mask and center the reference person. After this, we randomly apply geometric augmentations (scaling, rotation, and cutout). Color augmentations are important as, during training, the frames within the same video would usually have similar lighting and brightness. However, this may not be the case during inference, when we want to insert a random person in a random scene.

\vspace{-0.2em}
\subsection{Implementation details}
\vspace{-0.2em}

We train all models at $256\times256$ resolution. We encode these images using an autoencoder to a latent space of $32\times32\times4$ ($8\times$ downsample) resolution. The denoising backbone is based on time-conditional UNet~\cite{ronneberger2015u}. Following prior diffusion inpainting works~\cite{saharia2022palette, rombach2022high}, we concatenate the noisy image with the mask and the masked image. We pass the reference person through an image encoder and use the resulting features to condition the UNet via cross-attention. The mask and the masked image are concatenated as they are spatially aligned with the final output, whereas the reference person is injected through cross-attention as it would not be aligned due to having a different pose. We present ablations of different image encoders in our experiments. We also initialize our model with weights from Stable Diffusion's checkpoint~\cite{rombach2022high}.

At training time, to encourage better quality for the human hallucination task, we drop the person-conditioning 10\% of the time. We also drop both masked image and person-conditioning 10\% of the time to learn the full unconditional distribution and support classifier-free guidance. At test time, we use the DDIM sampler~\cite{song2021ddim} for 200 steps for all our results.

\label{sec:model-details}

\vspace{-0.2em}
\section{Experiments}
\label{sec:experiments}
\vspace{-0.2em}

We present evaluations on a few different tasks. First, we show results on conditional generation with a reference person in Sec.~\ref{sec:exp-conditional}. We also present ablations of data, architecture, and CFG in this section. We then present results on person hallucinations in Sec.~\ref{sec:exp-human-hall} and scene hallucinations in Sec.~\ref{sec:exp-scene-hall} and compare with Stable Diffusion~\cite{rombach2022high} and DALL-E 2~\cite{ramesh2022hierarchical} as baselines. We present additional results in the supp.\ material along with a discussion of failure cases.

\xhdr{Metrics.} We primarily use two quantitative metrics. First is FID (Fr\'echet Inception Distance)~\cite{fidscore}, which measures realism by comparing the distributions of Inception~\cite{szegedy2015going} network features of generated images with real images. We measure FID on 50K images, unless specified as FID-10K, wherein we use 10K images. Second is PCKh~\cite{andriluka20142d}, which measures accurate human positioning by computing the percentage of correct pose keypoints (within a radius relative to the head size). We use OpenPose~\cite{cao2019openpose} to detect keypoints of generated and real images.

\begin{figure*}

\centering
    \begin{subfigure}{\linewidth}
    \centering
        \begin{minipage}{\linewidth}
        \includegraphics[width=.105\linewidth]{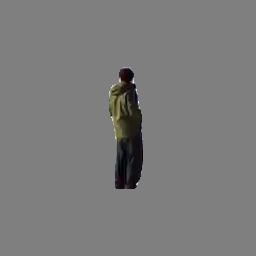}
        \includegraphics[width=.105\linewidth]{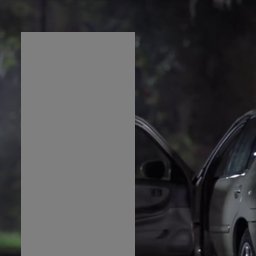}
        \hspace{-0.4em}
        \includegraphics[width=.105\linewidth]{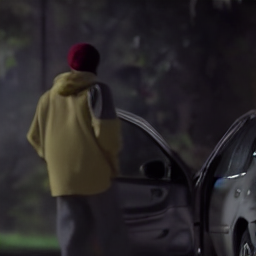}
        \includegraphics[width=.105\linewidth]{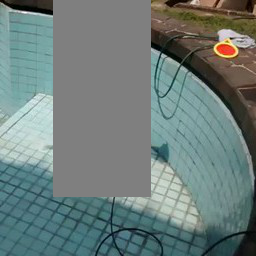}
        \hspace{-0.4em}
        \includegraphics[width=.105\linewidth]{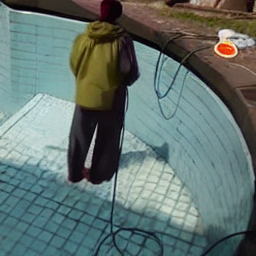}
        \includegraphics[width=.105\linewidth]{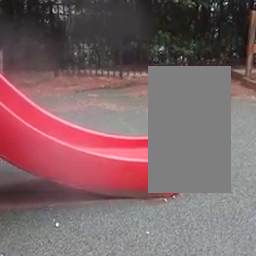}
        \hspace{-0.4em}
        \includegraphics[width=.105\linewidth]{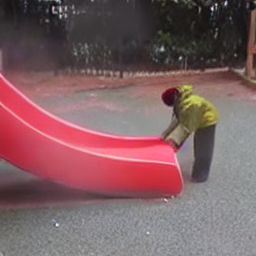}
        \includegraphics[width=.105\linewidth]{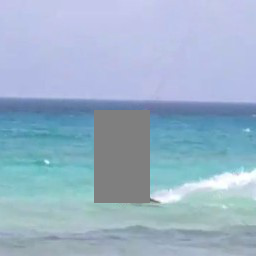}
        \hspace{-0.4em}
        \includegraphics[width=.105\linewidth]{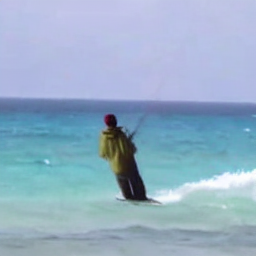}
        \end{minipage}
    \end{subfigure}
    \begin{subfigure}{\linewidth}
    \centering
        \begin{minipage}{\linewidth}
        \includegraphics[width=.105\linewidth]{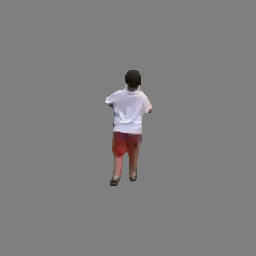}
        \includegraphics[width=.105\linewidth]{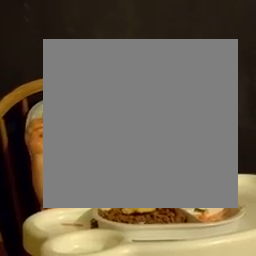}
        \hspace{-0.4em}
        \includegraphics[width=.105\linewidth]{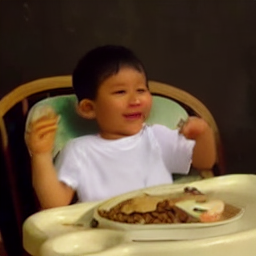}
        \includegraphics[width=.105\linewidth]{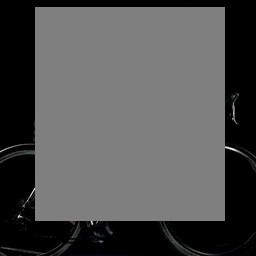}
        \hspace{-0.4em}
        \includegraphics[width=.105\linewidth]{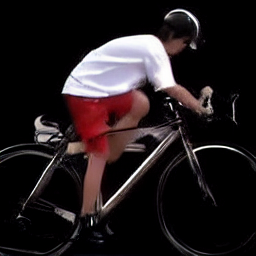}
        \includegraphics[width=.105\linewidth]{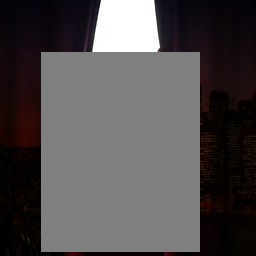}
        \hspace{-0.4em}
        \includegraphics[width=.105\linewidth]{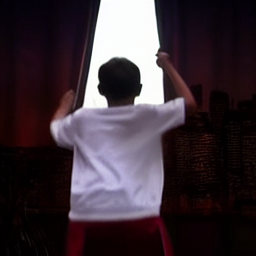}
        \includegraphics[width=.105\linewidth]{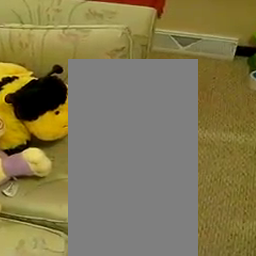}
        \hspace{-0.4em}
        \includegraphics[width=.105\linewidth]{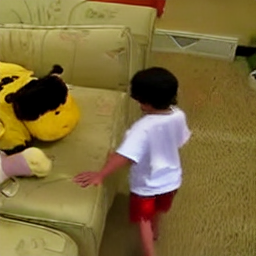}
        \end{minipage}
    \end{subfigure}
    \begin{subfigure}{\linewidth}
    \centering
        \begin{minipage}{\linewidth}
        \includegraphics[width=.105\linewidth]{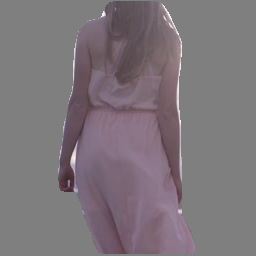}
        \includegraphics[width=.105\linewidth]{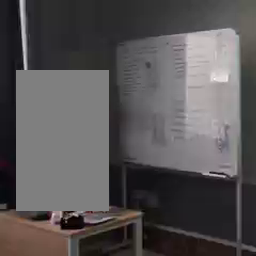}
        \hspace{-0.4em}
        \includegraphics[width=.105\linewidth]{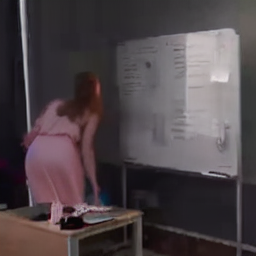}
        \includegraphics[width=.105\linewidth]{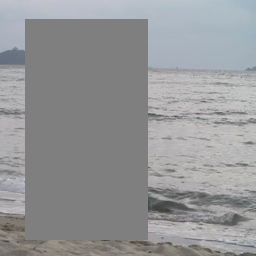}
        \hspace{-0.4em}
        \includegraphics[width=.105\linewidth]{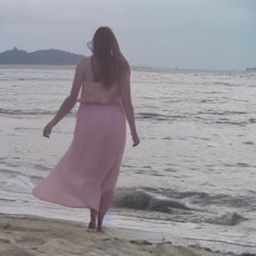}
        \includegraphics[width=.105\linewidth]{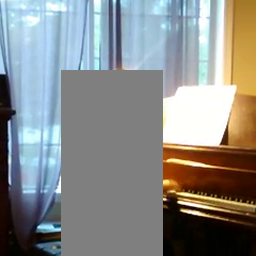}
        \hspace{-0.4em}
        \includegraphics[width=.105\linewidth]{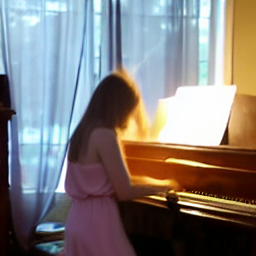}
        \includegraphics[width=.105\linewidth]{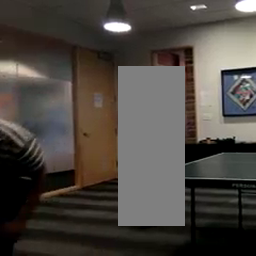}
        \hspace{-0.4em}
        \includegraphics[width=.105\linewidth]{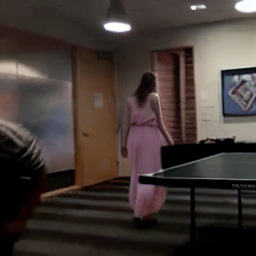}
        \end{minipage}
    \end{subfigure}
    \begin{subfigure}{\linewidth}
    \centering
        \begin{minipage}{\linewidth}
        \includegraphics[width=.105\linewidth]{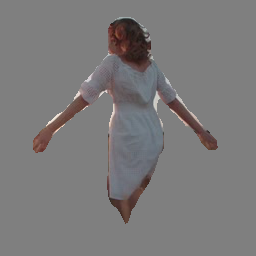}
        \includegraphics[width=.105\linewidth]{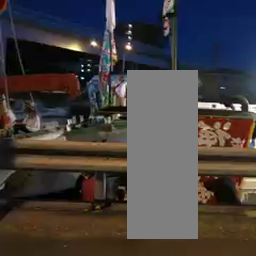}
        \hspace{-0.4em}
        \includegraphics[width=.105\linewidth]{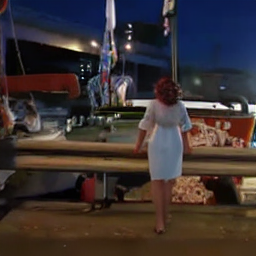}
        \includegraphics[width=.105\linewidth]{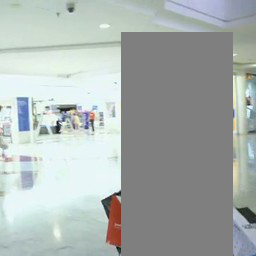}
        \hspace{-0.4em}
        \includegraphics[width=.105\linewidth]{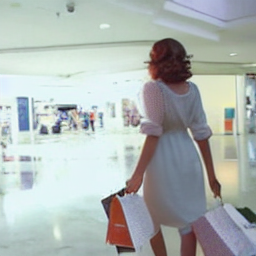}
        \includegraphics[width=.105\linewidth]{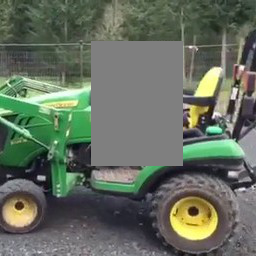}
        \hspace{-0.4em}
        \includegraphics[width=.105\linewidth]{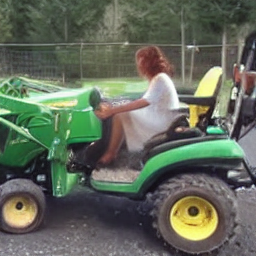}
        \includegraphics[width=.105\linewidth]{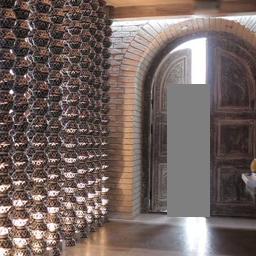}
        \hspace{-0.4em}
        \includegraphics[width=.105\linewidth]{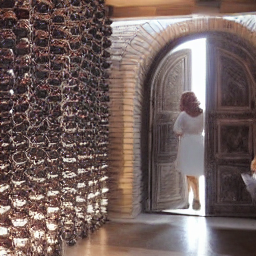}
        \end{minipage}
    \end{subfigure}
    
    \begin{subfigure}{\linewidth}
    \centering
        \begin{minipage}{\linewidth}
        \includegraphics[width=.105\linewidth]{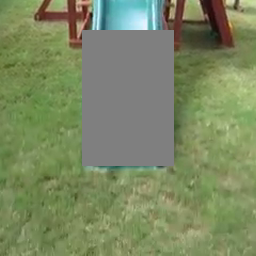}
        \includegraphics[width=.105\linewidth]{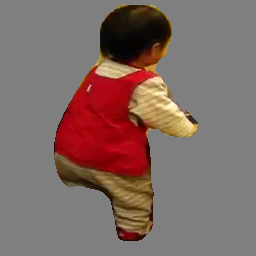}
        \hspace{-0.4em}
        \includegraphics[width=.105\linewidth]{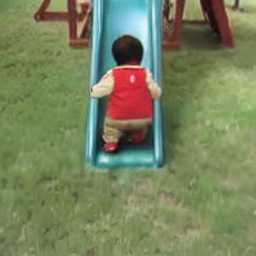}
        \includegraphics[width=.105\linewidth]{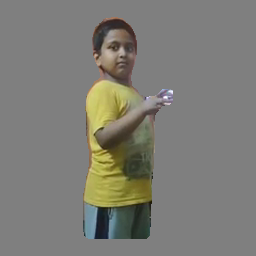}
        \hspace{-0.4em}
        \includegraphics[width=.105\linewidth]{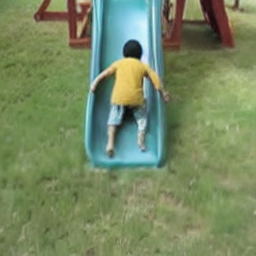}
        \includegraphics[width=.105\linewidth]{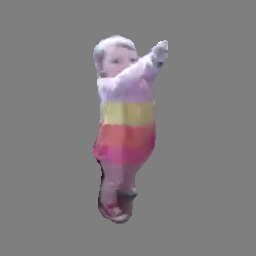}
        \hspace{-0.4em}
        \includegraphics[width=.105\linewidth]{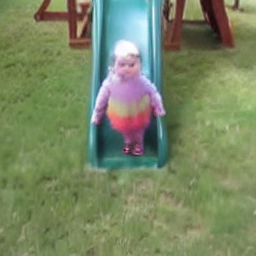}
        \includegraphics[width=.105\linewidth]{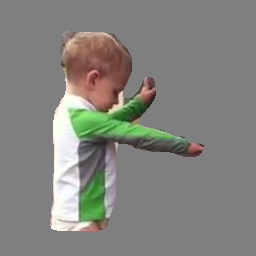}
        \hspace{-0.4em}
        \includegraphics[width=.105\linewidth]{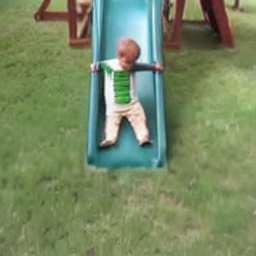}
        \end{minipage}
    \end{subfigure}
    \begin{subfigure}{\linewidth}
    \centering
        \begin{minipage}{\linewidth}
        \includegraphics[width=.105\linewidth]{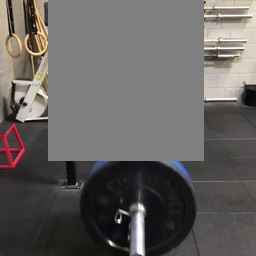}
        \includegraphics[width=.105\linewidth]{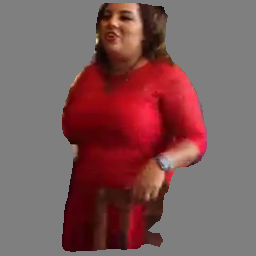}
        \hspace{-0.4em}
        \includegraphics[width=.105\linewidth]{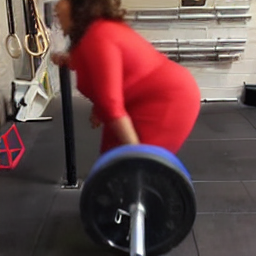}
        \includegraphics[width=.105\linewidth]{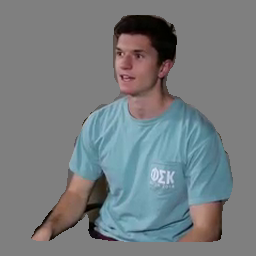}
        \hspace{-0.4em}
        \includegraphics[width=.105\linewidth]{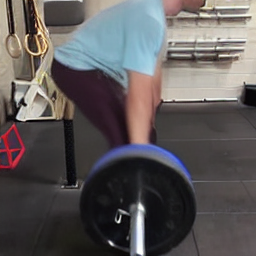}
        \includegraphics[width=.105\linewidth]{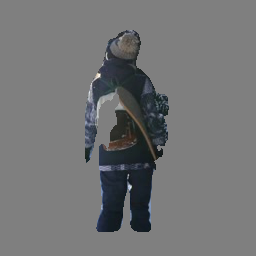}
        \hspace{-0.4em}
        \includegraphics[width=.105\linewidth]{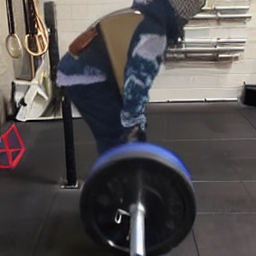}
        \includegraphics[width=.105\linewidth]{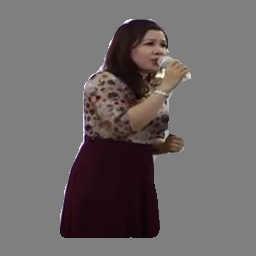}
        \hspace{-0.4em}
        \includegraphics[width=.105\linewidth]{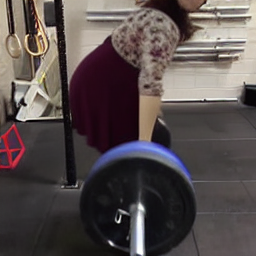}
        \end{minipage}
    \end{subfigure}
    \begin{subfigure}{\linewidth}
    \centering
        \begin{minipage}{\linewidth}
        \includegraphics[width=.105\linewidth]{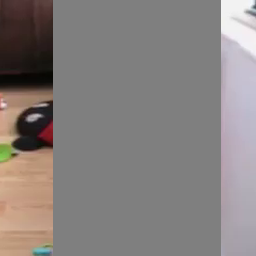}
        \includegraphics[width=.105\linewidth]{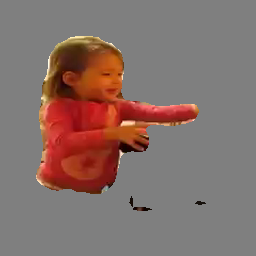}
        \hspace{-0.4em}
        \includegraphics[width=.105\linewidth]{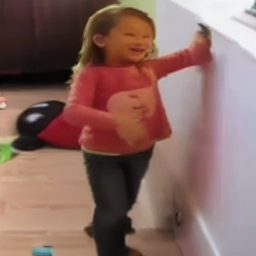}
        \includegraphics[width=.105\linewidth]{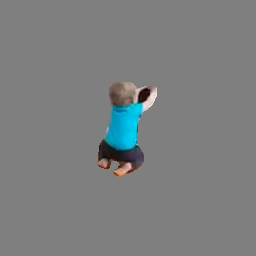}
        \hspace{-0.4em}
        \includegraphics[width=.105\linewidth]{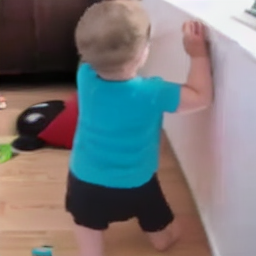}
        \includegraphics[width=.105\linewidth]{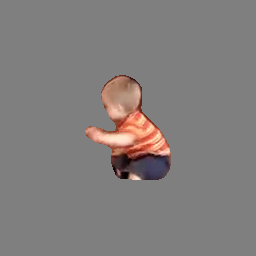}
        \hspace{-0.4em}
        \includegraphics[width=.105\linewidth]{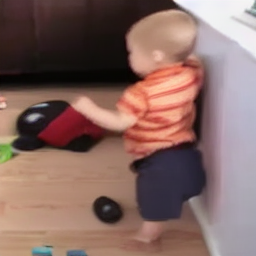}
        \includegraphics[width=.105\linewidth]{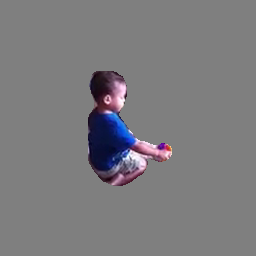}
        \hspace{-0.4em}
        \includegraphics[width=.105\linewidth]{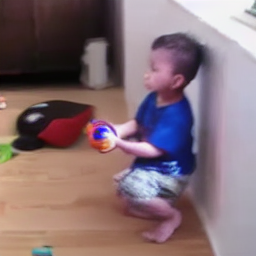}
        \end{minipage}
    \end{subfigure}
    \begin{subfigure}{\linewidth}
    \centering
        \begin{minipage}{\linewidth}
        \includegraphics[width=.105\linewidth]{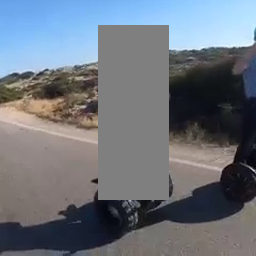}
        \includegraphics[width=.105\linewidth]{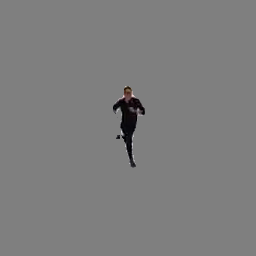}
        \hspace{-0.4em}
        \includegraphics[width=.105\linewidth]{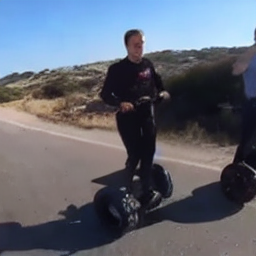}
        \includegraphics[width=.105\linewidth]{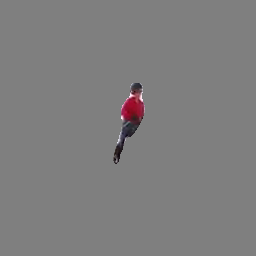}
        \hspace{-0.4em}
        \includegraphics[width=.105\linewidth]{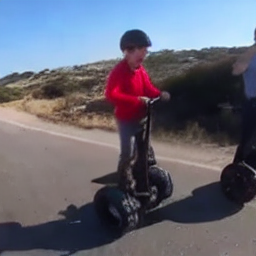}
        \includegraphics[width=.105\linewidth]{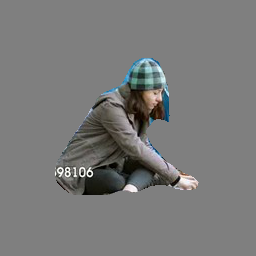}
        \hspace{-0.4em}
        \includegraphics[width=.105\linewidth]{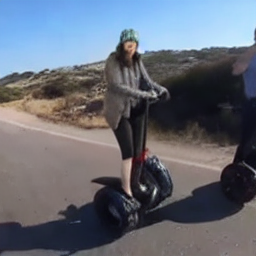}
        \includegraphics[width=.105\linewidth]{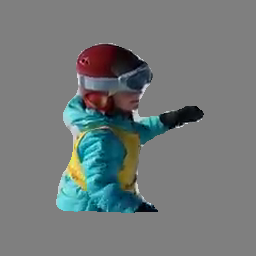}
        \hspace{-0.4em}
        \includegraphics[width=.105\linewidth]{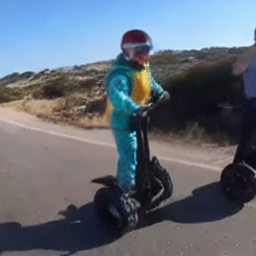}
        \end{minipage}
    \end{subfigure}

\caption{
\textbf{Qualitative results of conditional generation.} In the top 4 rows, we show a reference person in the first column, followed by four pairs of masked scene image and corresponding result.  In the bottom 4 rows, we show a masked scene image in the first column, followed by four pairs of reference person and corresponding result. Our results have the reference person re-posed correctly according to the scene.  
}
\label{fig:res-cond-fig}
\end{figure*}

\vspace{-0.2em}
\subsection{Conditional generation}
\label{sec:exp-conditional}
\vspace{-0.2em}

We evaluate the conditional task of generating a target image given a masked scene image and a reference person.

All our models were trained on 32 A100s for 100K steps with a batch size of 1024. We compute the metrics on the held-out set of 50K videos, by trying to inpaint the first masked frame for each video. We choose a reference person from a different video to make the task challenging and use the same mapping for all evaluations.

We present three sets of ablations. 
\textbf{Data.} We experiment with using different training data. We simulate image-only supervision by taking the masked scene image and the reference person from the same frame. We also ablate with data augmentations turned on and off.  
\textbf{Encoders.} We experiment with using the first-stage VAE features, passed in as concatenation instead of CLIP ViT-L/14 embeddings. 
\textbf{UNet.} We experiment with a smaller UNet (430M) compared to ours (860M). We also study the effects of initializing with a pre-trained checkpoint.

\begin{table}
\caption{Comparison of metrics for different ablations. First set are on data used for training, second set are on encoders and the final set are on model scaling and effects of pretraining. Metrics used are FID (lower is better) and PCKh (higher is better). 
}
\vspace{-1.0em}
\small
\centering\vspace{5pt}
\begin{tabular}{lcc}

\toprule

Method & FID $\downarrow$ & PCKh $\uparrow$ \\

\midrule

Image (w/o aug) & 13.174 & 8.321\\
Image (w/ aug) & 13.008 & 10.660\\
Video (w/o aug) & 12.103 & 15.797 \\

\midrule

VAE KL-8x (concat) & 14.956 & 13.020 \\

\midrule

Small (400M, scratch) & 12.366 & 15.095 \\
Large (scratch)  & 11.232 & 15.873 \\
Large (fine-tune) & 10.078 & 17.602  \\

\bottomrule
\end{tabular}
\label{tab:ablations}
\vspace{-1.0em}
\end{table}

\begin{figure}
\centering

\begin{subfigure}{.48\linewidth}
  \centering
  \includegraphics[width=\linewidth]{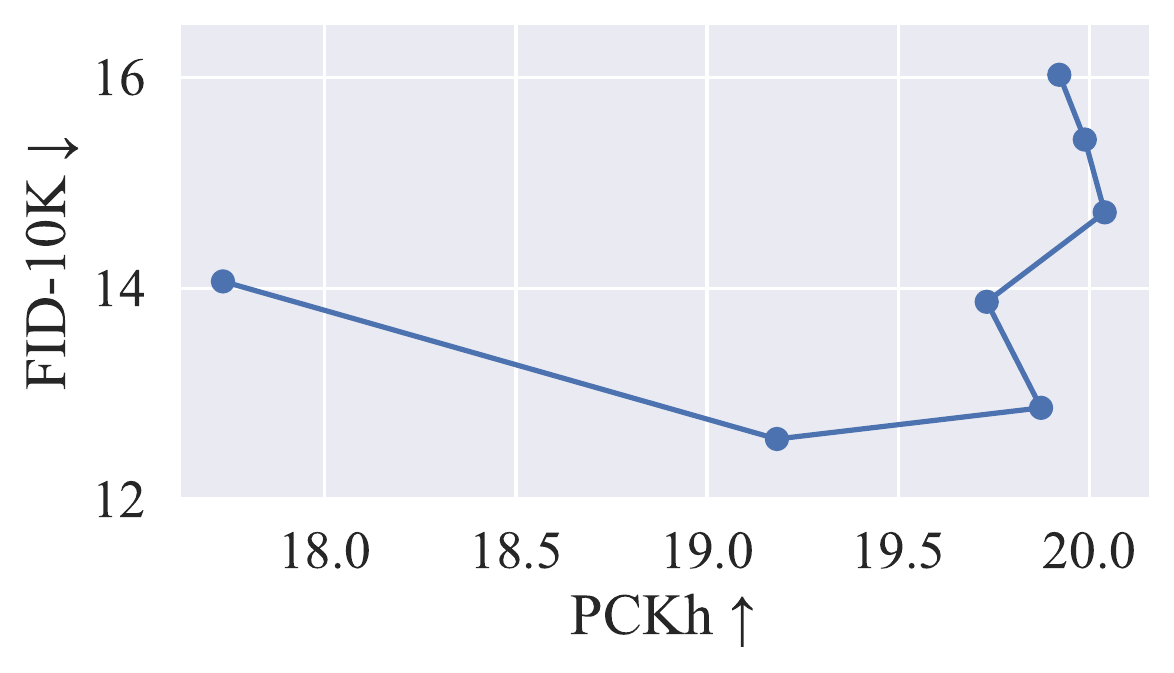}
  \caption{Person conditioning}
  \label{fig:cfg-person-conditioning}
\end{subfigure}
\begin{subfigure}{.48\linewidth}
  \centering
  \includegraphics[width=\linewidth]{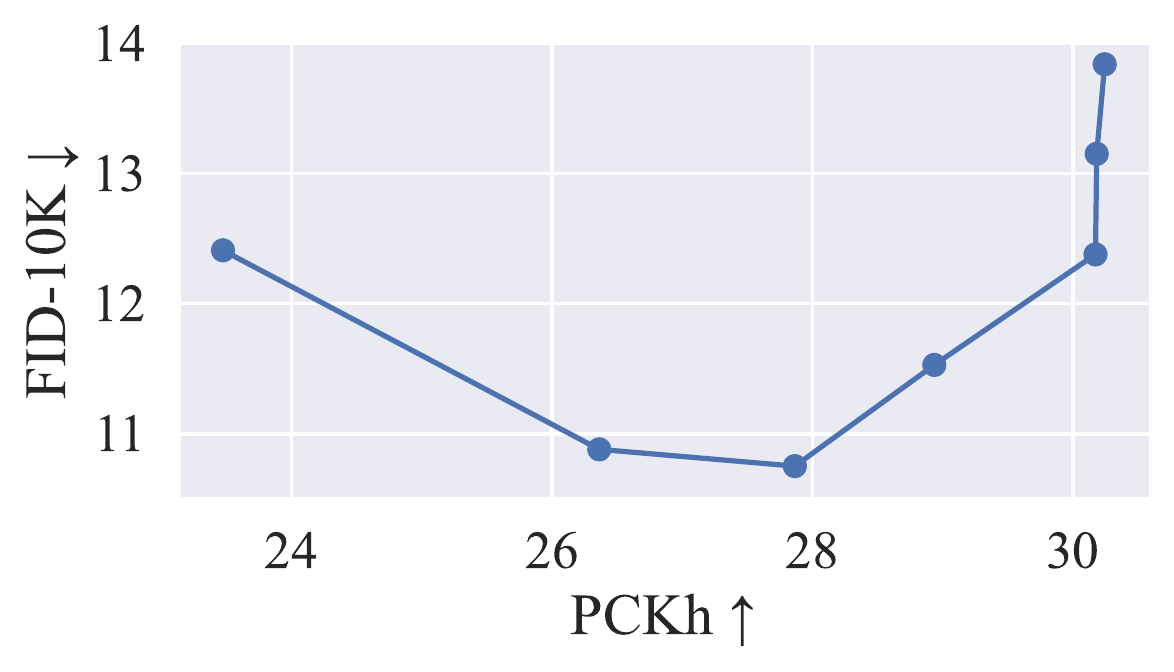}
  \caption{Person hallucination}
  \label{fig:cfg-person-hallucination}
\end{subfigure}

\vspace{-0.6em}
\caption{\textbf{Classifier-free guidance.} Effect of increasing CFG guidance scale. Evaluated at [1.0, 1.5, 2.0, 3.0, 4.0, 5.0, 6.0].
}
\vspace{-1.4em}
\label{fig:cfg-u-plot}
\end{figure}

\begin{figure}[!htb]
\centering

    \begin{subfigure}{\linewidth}
    \centering
        \begin{minipage}{\linewidth}
        \includegraphics[width=.195\linewidth]{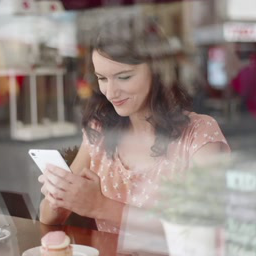}
        \hspace{-0.4em}
        \includegraphics[width=.195\linewidth]{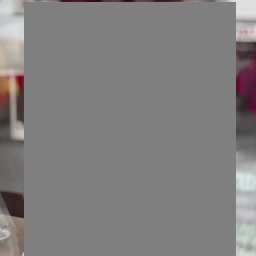}
        \hspace{-0.4em}
        \includegraphics[width=.195\linewidth]{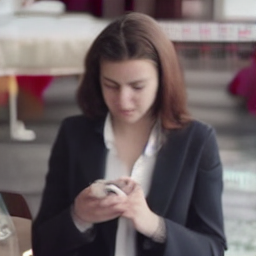}
        \hspace{-0.4em}
        \includegraphics[width=.195\linewidth]{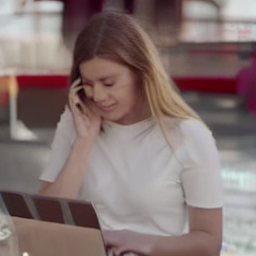}
        \hspace{-0.4em}
        \includegraphics[width=.195\linewidth]{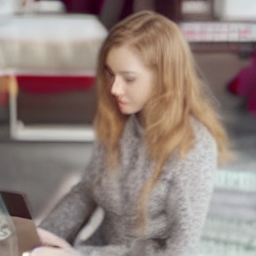}
        \end{minipage}
    \end{subfigure}

    \begin{subfigure}{\linewidth}
    \centering
        \begin{minipage}{\linewidth}
        \includegraphics[width=.195\linewidth]{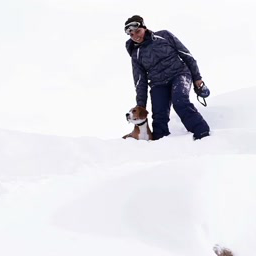}
        \hspace{-0.4em}
        \includegraphics[width=.195\linewidth]{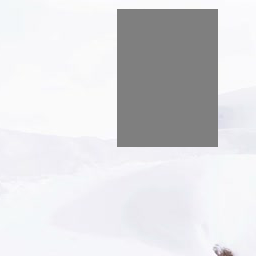}
        \hspace{-0.4em}
        \includegraphics[width=.195\linewidth]{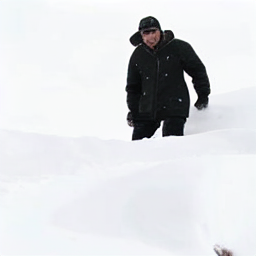}
        \hspace{-0.4em}
        \includegraphics[width=.195\linewidth]{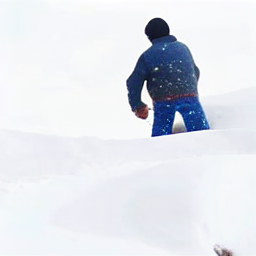}
        \hspace{-0.4em}
        \includegraphics[width=.195\linewidth]{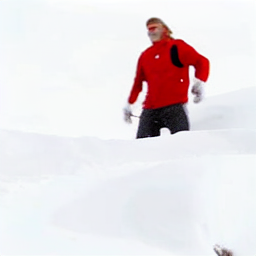}
        \end{minipage}
    \end{subfigure}

    \begin{subfigure}{\linewidth}
    \centering
        \begin{minipage}{\linewidth}
        \includegraphics[width=.195\linewidth]{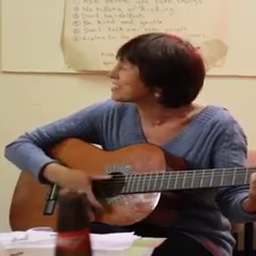}
        \hspace{-0.4em}
        \includegraphics[width=.195\linewidth]{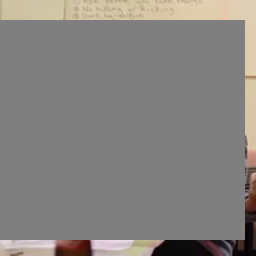}
        \hspace{-0.4em}
        \includegraphics[width=.195\linewidth]{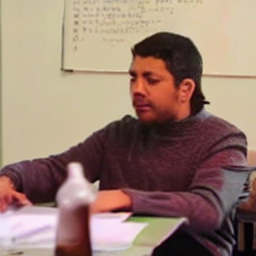}
        \hspace{-0.4em}
        \includegraphics[width=.195\linewidth]{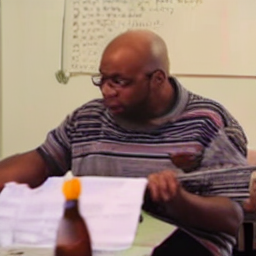}
        \hspace{-0.4em}
        \includegraphics[width=.195\linewidth]{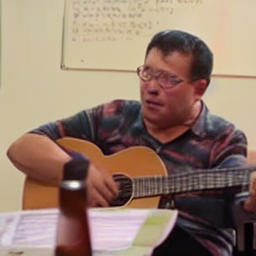}
        \end{minipage}
    \end{subfigure}
    \begin{subfigure}{\linewidth}
    \centering
        \begin{minipage}{\linewidth}
        \includegraphics[width=.195\linewidth]{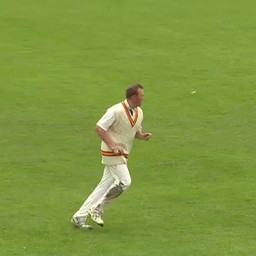}
        \hspace{-0.4em}
        \includegraphics[width=.195\linewidth]{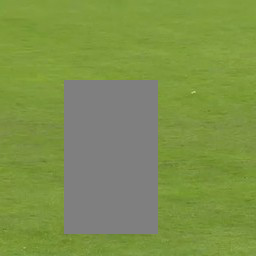}
        \hspace{-0.4em}
        \includegraphics[width=.195\linewidth]{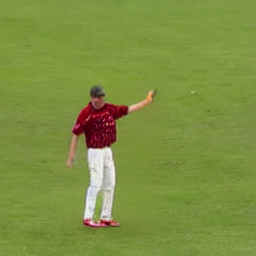}
        \hspace{-0.4em}
        \includegraphics[width=.195\linewidth]{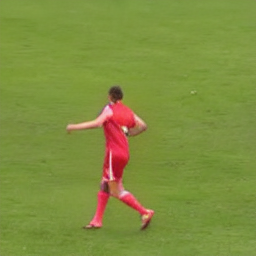}
        \hspace{-0.4em}
        \includegraphics[width=.195\linewidth]{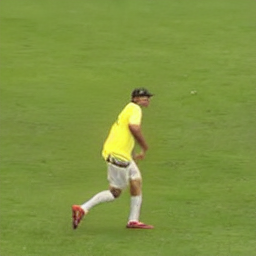}
        \end{minipage}
    \end{subfigure}
    \begin{subfigure}{\linewidth}
    \centering
        \begin{minipage}{\linewidth}
        \includegraphics[width=.195\linewidth]{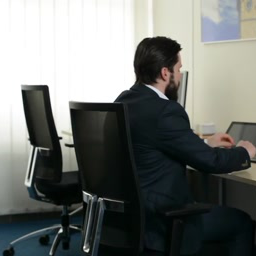}
        \hspace{-0.4em}
        \includegraphics[width=.195\linewidth]{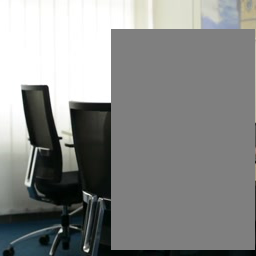}
        \hspace{-0.4em}
        \includegraphics[width=.195\linewidth]{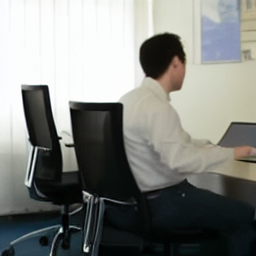}
        \hspace{-0.4em}
        \includegraphics[width=.195\linewidth]{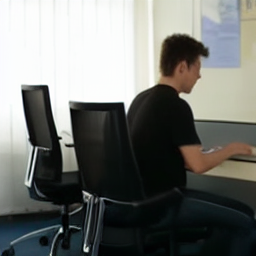}
        \hspace{-0.4em}
        \includegraphics[width=.195\linewidth]{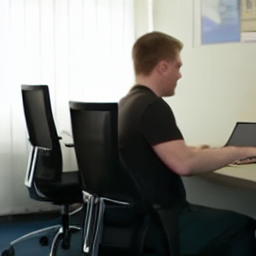}
        \end{minipage}
    \end{subfigure}

    \begin{subfigure}{\linewidth}
    \centering
        \begin{minipage}{\linewidth}
        \includegraphics[width=.195\linewidth]{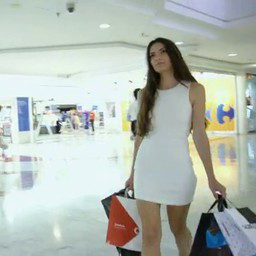}
        \hspace{-0.4em}
        \includegraphics[width=.195\linewidth]{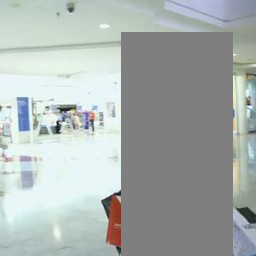}
        \hspace{-0.4em}
        \includegraphics[width=.195\linewidth]{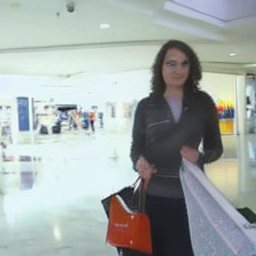}
        \hspace{-0.4em}
        \includegraphics[width=.195\linewidth]{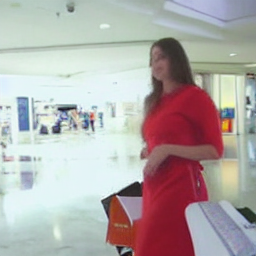}
        \hspace{-0.4em}
        \includegraphics[width=.195\linewidth]{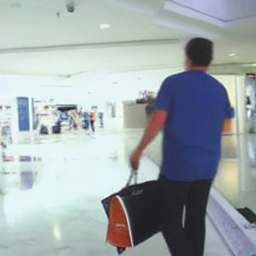}
        \end{minipage}
    \end{subfigure}

    \begin{subfigure}{\linewidth}
    \centering
        \begin{minipage}{\linewidth}
        \includegraphics[width=.195\linewidth]{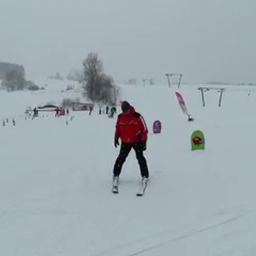}
        \hspace{-0.4em}
        \includegraphics[width=.195\linewidth]{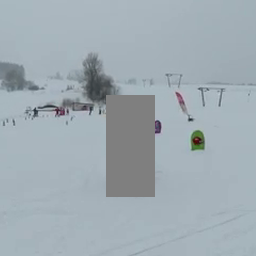}
        \hspace{-0.4em}
        \includegraphics[width=.195\linewidth]{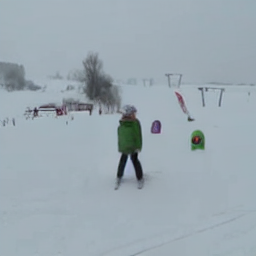}
        \hspace{-0.4em}
        \includegraphics[width=.195\linewidth]{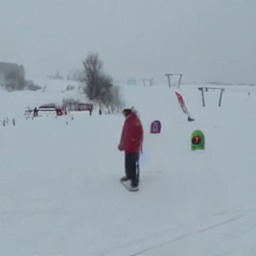}
        \hspace{-0.4em}
        \includegraphics[width=.195\linewidth]{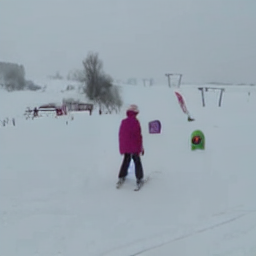}
        \end{minipage}
    \end{subfigure}

    \begin{subfigure}{\linewidth}
    \centering
        \begin{minipage}{\linewidth}
        \includegraphics[width=.195\linewidth]{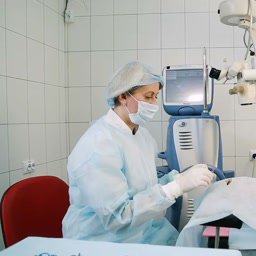}
        \hspace{-0.4em}
        \includegraphics[width=.195\linewidth]{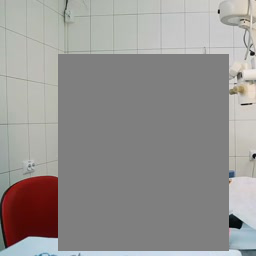}
        \hspace{-0.4em}
        \includegraphics[width=.195\linewidth]{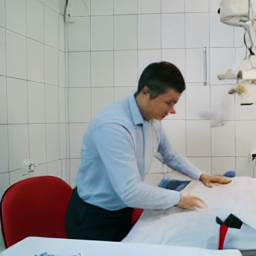}
        \hspace{-0.4em}
        \includegraphics[width=.195\linewidth]{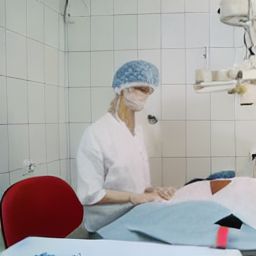}
        \hspace{-0.4em}
        \includegraphics[width=.195\linewidth]{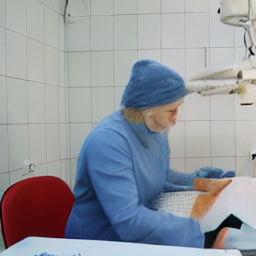}
        \end{minipage}
    \end{subfigure}
    \begin{subfigure}{\linewidth}
    \centering
        \begin{minipage}{\linewidth}
        \includegraphics[width=.195\linewidth]{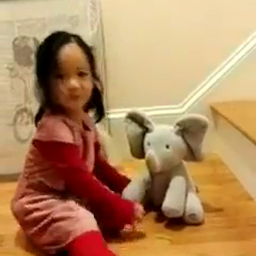}
        \hspace{-0.4em}
        \includegraphics[width=.195\linewidth]{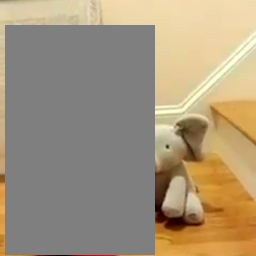}
        \hspace{-0.4em}
        \includegraphics[width=.195\linewidth]{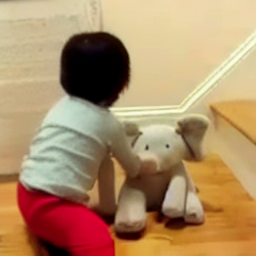}
        \hspace{-0.4em}
        \includegraphics[width=.195\linewidth]{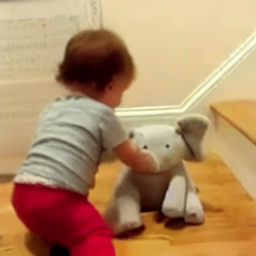}
        \hspace{-0.4em}
        \includegraphics[width=.195\linewidth]{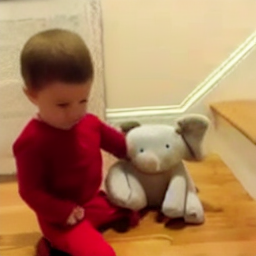}
        \end{minipage}
    \end{subfigure}

\vspace{-0.5em}
\caption{
\textbf{Qualitative results of person hallucination.} From left to right, groundtruth image, masked scene image, 3 hallucinated persons in the scene. Our method can hallucinate plausible pose and appearance.
}
\vspace{-1.5em}
\label{fig:res-person-hall}
\end{figure}

\begin{figure}
\centering
    \begin{subfigure}{\linewidth}
    \centering
        \begin{minipage}{\linewidth}
        \includegraphics[width=.195\linewidth]{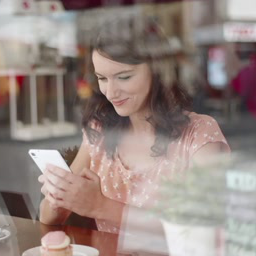}
        \hspace{-0.4em}
        \includegraphics[width=.195\linewidth]{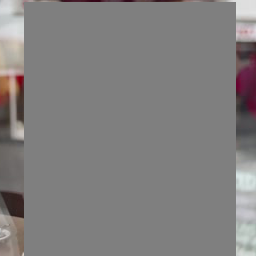}
        \hspace{-0.4em}
        \includegraphics[width=.195\linewidth]{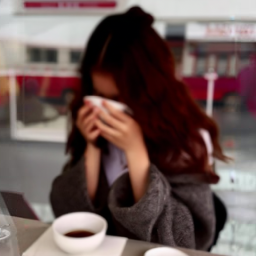}
        \hspace{-0.4em}
        \includegraphics[width=.195\linewidth]{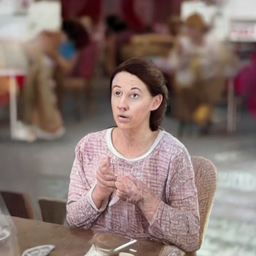}
        \hspace{-0.4em}
        \includegraphics[width=.195\linewidth]{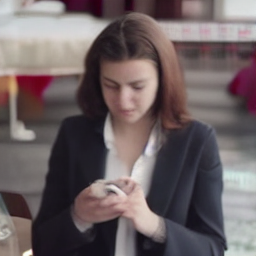}
        \end{minipage}
    \end{subfigure}
    \begin{subfigure}{\linewidth}
    \centering
        \begin{minipage}{\linewidth}
        \includegraphics[width=.195\linewidth]{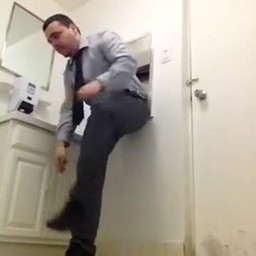}
        \hspace{-0.4em}
        \includegraphics[width=.195\linewidth]{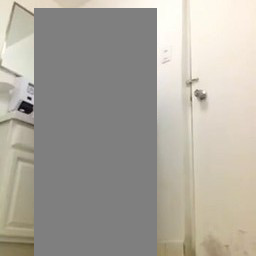}
        \hspace{-0.4em}
        \includegraphics[width=.195\linewidth]{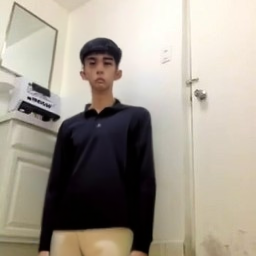}
        \hspace{-0.4em}
        \includegraphics[width=.195\linewidth]{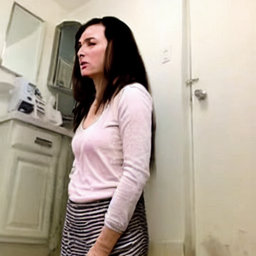}
        \hspace{-0.4em}
        \includegraphics[width=.195\linewidth]{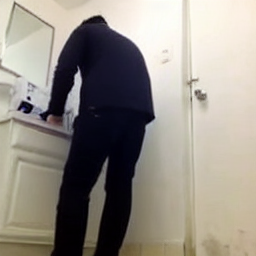}
        \end{minipage}
    \end{subfigure}
    \begin{subfigure}{\linewidth}
    \centering
        \begin{minipage}{\linewidth}
        \includegraphics[width=.195\linewidth]{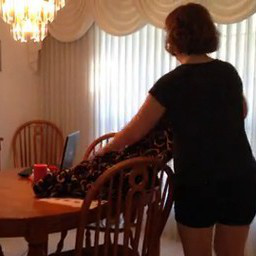}
        \hspace{-0.4em}
        \includegraphics[width=.195\linewidth]{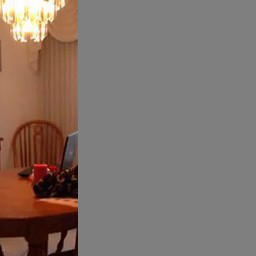}
        \hspace{-0.4em}
        \includegraphics[width=.195\linewidth]{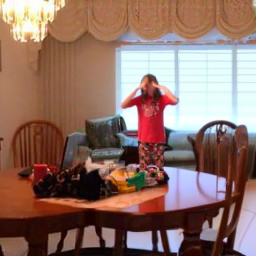}
        \hspace{-0.4em}
        \includegraphics[width=.195\linewidth]{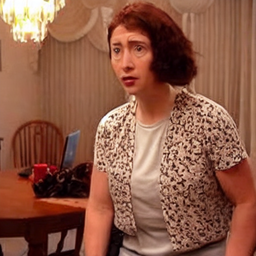}
        \hspace{-0.4em}
        \includegraphics[width=.195\linewidth]{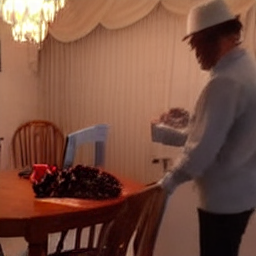}
        \end{minipage}
    \end{subfigure}
    \begin{subfigure}{\linewidth}
    \centering
        \begin{minipage}{\linewidth}
        \includegraphics[width=.195\linewidth]{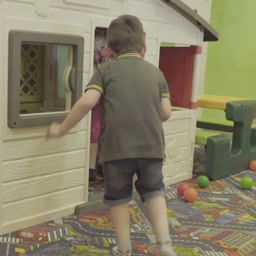}
        \hspace{-0.4em}
        \includegraphics[width=.195\linewidth]{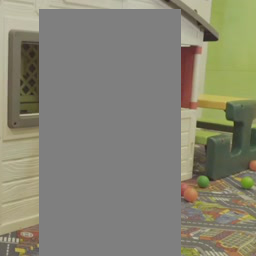}
        \hspace{-0.4em}
        \includegraphics[width=.195\linewidth]{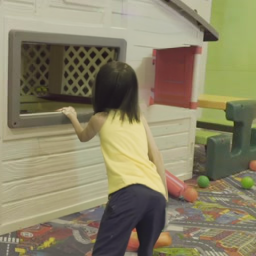}
        \hspace{-0.4em}
        \includegraphics[width=.195\linewidth]{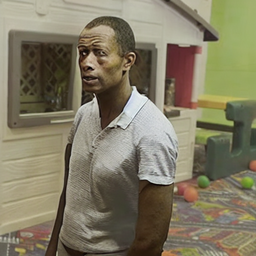}
        \hspace{-0.4em}
        \includegraphics[width=.195\linewidth]{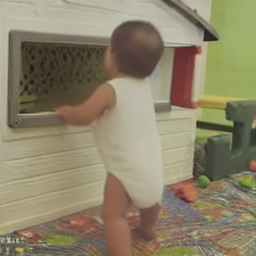}
        \end{minipage}
    \end{subfigure}
    \begin{subfigure}{\linewidth}
    \centering
        \begin{minipage}{\linewidth}
        \includegraphics[width=.195\linewidth]{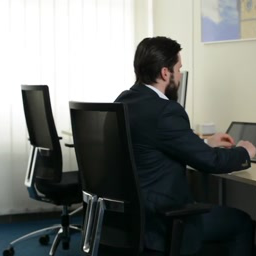}
        \hspace{-0.4em}
        \includegraphics[width=.195\linewidth]{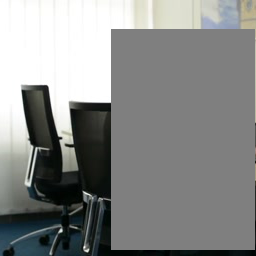}
        \hspace{-0.4em}
        \includegraphics[width=.195\linewidth]{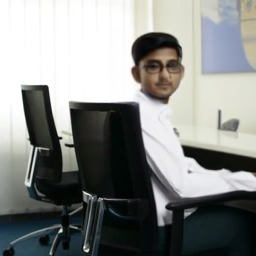}
        \hspace{-0.4em}
        \includegraphics[width=.195\linewidth]{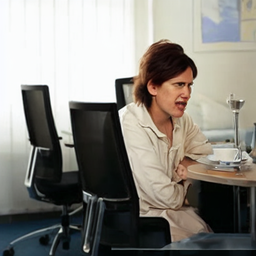}
        \hspace{-0.4em}
        \includegraphics[width=.195\linewidth]{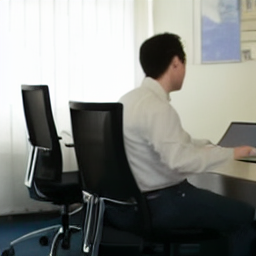}
        \end{minipage}
    \end{subfigure}

\vspace{-0.5em}
\caption{
\textbf{Baseline comparisons for person hallucination.} From left to right, ground-truth, masked scene image, DALL-E 2 result, Stable Diffusion result and our result. Our model does the best job in hallucinating humans consistent with the context.  
}
\label{fig:person-hall-baselines}
\vspace{-0.5em}
\end{figure}

\begin{figure}
\centering
    \begin{subfigure}{\linewidth}
    \centering
        \begin{minipage}{\linewidth}
        \includegraphics[width=.195\linewidth]{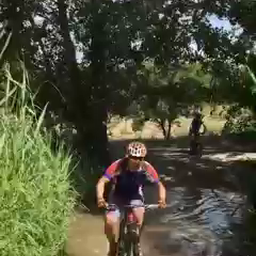}
        \hspace{-0.4em}
        \includegraphics[width=.195\linewidth]{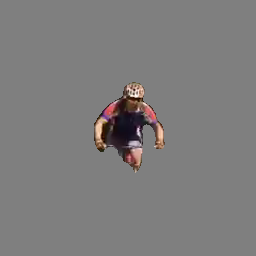}
        \hspace{-0.4em}
        \includegraphics[width=.195\linewidth]{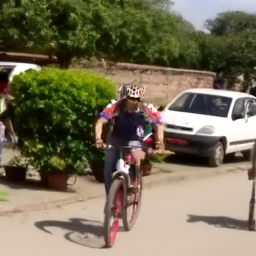}
        \hspace{-0.4em}
        \includegraphics[width=.195\linewidth]{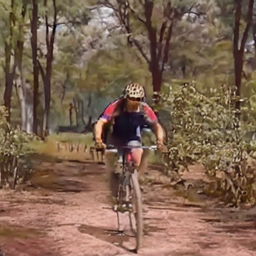}
        \hspace{-0.4em}
        \includegraphics[width=.195\linewidth]{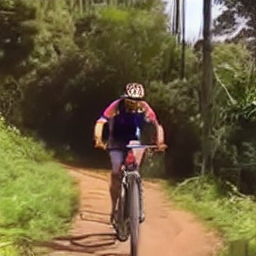}
        \end{minipage}
    \end{subfigure}
    \begin{subfigure}{\linewidth}
    \centering
        \begin{minipage}{\linewidth}
        \includegraphics[width=.195\linewidth]{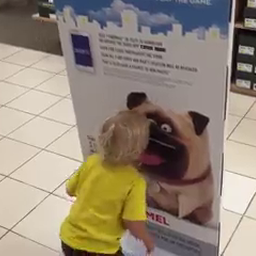}
        \hspace{-0.4em}
        \includegraphics[width=.195\linewidth]{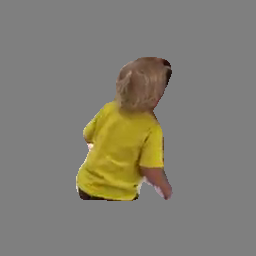}
        \hspace{-0.4em}
        \includegraphics[width=.195\linewidth]{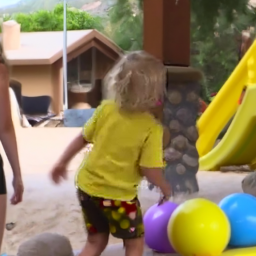}
        \hspace{-0.4em}
        \includegraphics[width=.195\linewidth]{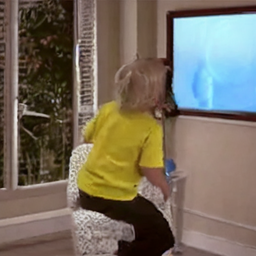}
        \hspace{-0.4em}
        \includegraphics[width=.195\linewidth]{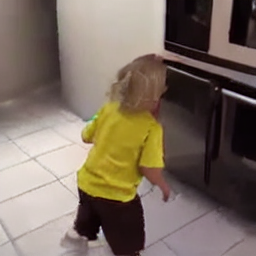}
        \end{minipage}
    \end{subfigure}
    \begin{subfigure}{\linewidth}
    \centering
        \begin{minipage}{\linewidth}
        \includegraphics[width=.195\linewidth]{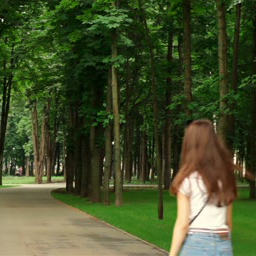}
        \hspace{-0.4em}
        \includegraphics[width=.195\linewidth]{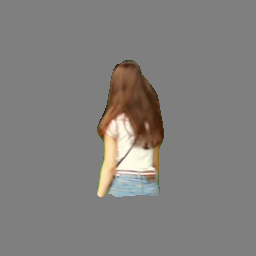}
        \hspace{-0.4em}
        \includegraphics[width=.195\linewidth]{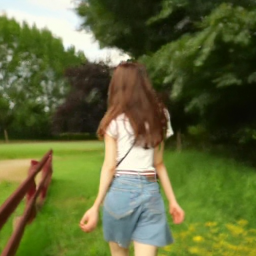}
        \hspace{-0.4em}
        \includegraphics[width=.195\linewidth]{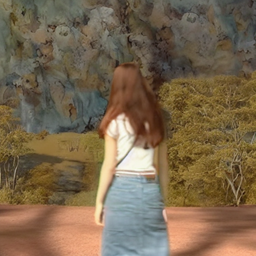}
        \hspace{-0.4em}
        \includegraphics[width=.195\linewidth]{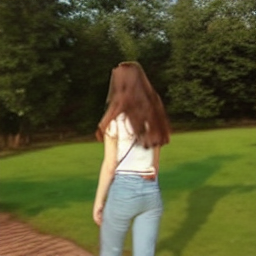}
        \end{minipage}
    \end{subfigure}
    \begin{subfigure}{\linewidth}
    \centering
        \begin{minipage}{\linewidth}
        \includegraphics[width=.195\linewidth]{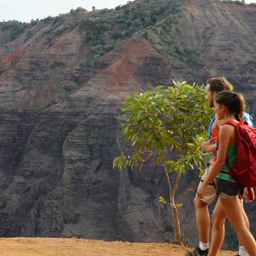}
        \hspace{-0.4em}
        \includegraphics[width=.195\linewidth]{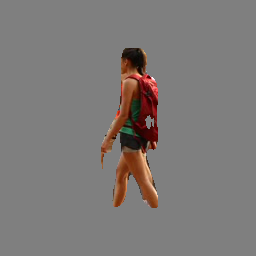}
        \hspace{-0.4em}
        \includegraphics[width=.195\linewidth]{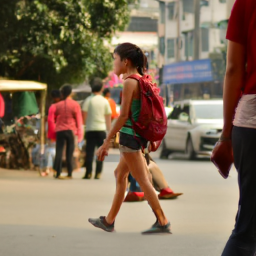}
        \hspace{-0.4em}
        \includegraphics[width=.195\linewidth]{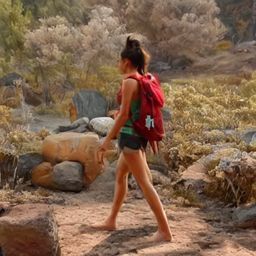}
        \hspace{-0.4em}
        \includegraphics[width=.195\linewidth]{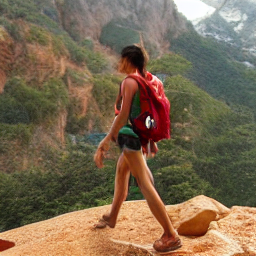}
        \end{minipage}
    \end{subfigure}
    \begin{subfigure}{\linewidth}
    \centering
        \begin{minipage}{\linewidth}
        \includegraphics[width=.195\linewidth]{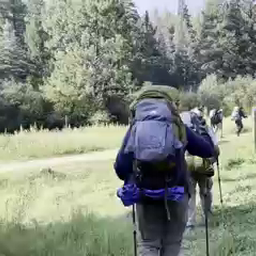}
        \hspace{-0.4em}
        \includegraphics[width=.195\linewidth]{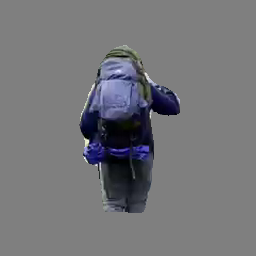}
        \hspace{-0.4em}
        \includegraphics[width=.195\linewidth]{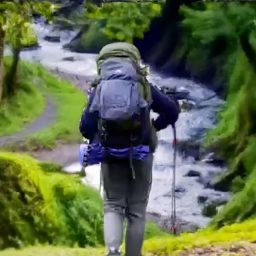}
        \hspace{-0.4em}
        \includegraphics[width=.195\linewidth]{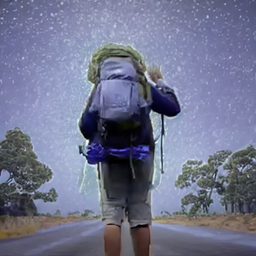}
        \hspace{-0.4em}
        \includegraphics[width=.195\linewidth]{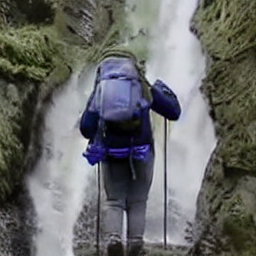}
        \end{minipage}
    \end{subfigure}

\vspace{-0.5em}
\caption{
\textbf{Baseline comparisons for scene hallucination.} From left to right, ground-truth, reference person, DALL-E 2 result, Stable Diffusion result, and our result. Our model does the best job of hallucinating the scene consistent with the reference person.
}
\label{fig:scene-hall-baselines}
\vspace{-1.5em}
\end{figure}

\begin{figure}[!htb]
\centering
    \begin{subfigure}{\linewidth}
    \centering
        \begin{minipage}{\linewidth}
        \includegraphics[width=.195\linewidth]{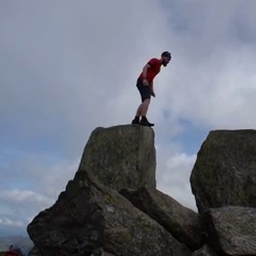}
        \hspace{-0.4em}
        \includegraphics[width=.195\linewidth]{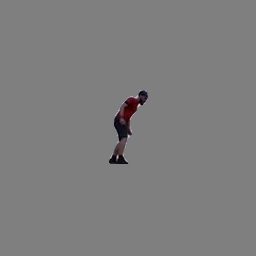}
        \hspace{-0.4em}
        \includegraphics[width=.195\linewidth]{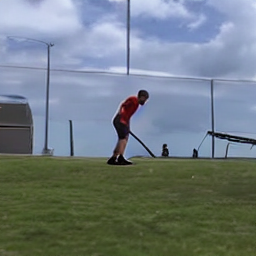}
        \hspace{-0.4em}
        \includegraphics[width=.195\linewidth]{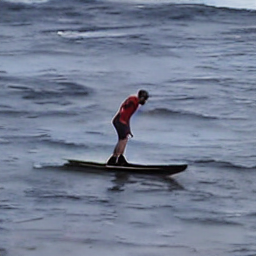}
        \hspace{-0.4em}
        \includegraphics[width=.195\linewidth]{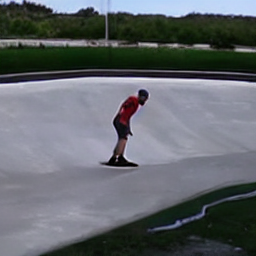}
        \end{minipage}
    \end{subfigure}
    \begin{subfigure}{\linewidth}
    \centering
        \begin{minipage}{\linewidth}
        \includegraphics[width=.195\linewidth]{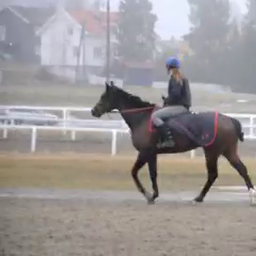}
        \hspace{-0.4em}
        \includegraphics[width=.195\linewidth]{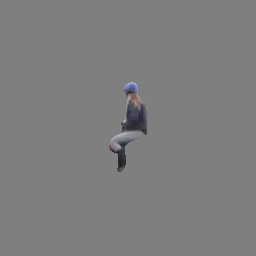}
        \hspace{-0.4em}
        \includegraphics[width=.195\linewidth]{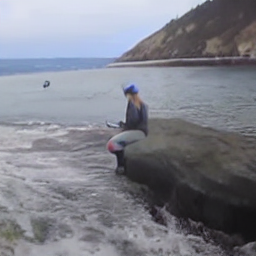}
        \hspace{-0.4em}
        \includegraphics[width=.195\linewidth]{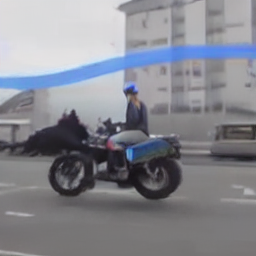}
        \hspace{-0.4em}
        \includegraphics[width=.195\linewidth]{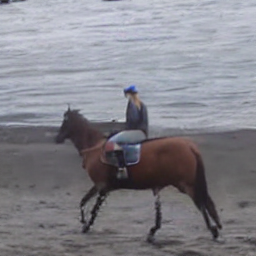}
        \end{minipage}
    \end{subfigure}
    \begin{subfigure}{\linewidth}
    \centering
        \begin{minipage}{\linewidth}
        \includegraphics[width=.195\linewidth]{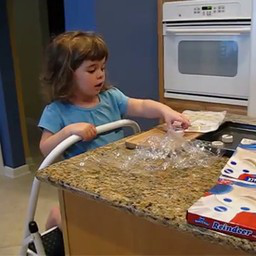}
        \hspace{-0.4em}
        \includegraphics[width=.195\linewidth]{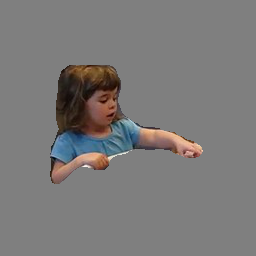}
        \hspace{-0.4em}
        \includegraphics[width=.195\linewidth]{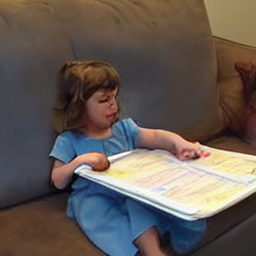}
        \hspace{-0.4em}
        \includegraphics[width=.195\linewidth]{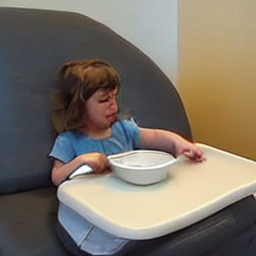}
        \hspace{-0.4em}
        \includegraphics[width=.195\linewidth]{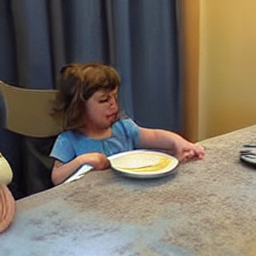}
        \end{minipage}
    \end{subfigure}
    \begin{subfigure}{\linewidth}
    \centering
        \begin{minipage}{\linewidth}
        \includegraphics[width=.195\linewidth]{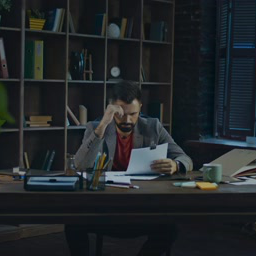}
        \hspace{-0.4em}
        \includegraphics[width=.195\linewidth]{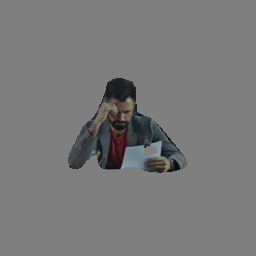}
        \hspace{-0.4em}
        \includegraphics[width=.195\linewidth]{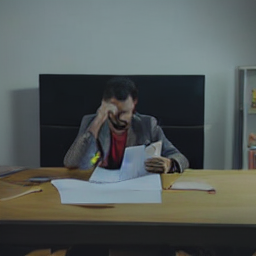}
        \hspace{-0.4em}
        \includegraphics[width=.195\linewidth]{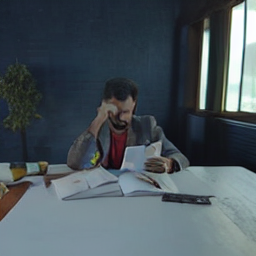}
        \hspace{-0.4em}
        \includegraphics[width=.195\linewidth]{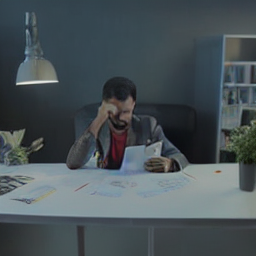}
        \end{minipage}
    \end{subfigure}

\caption{
\textbf{Constrained scene hallucination.} From left to right, ground-truth, reference person, 3 hallucinated scene samples where the pose and location of the person is constrained. Our hallucinated scene has consistent affordances with the reference person and the reference person stays unchanged.
}
\label{fig:scene-hall-center}
\end{figure}

\begin{figure}[!htb]
\centering
    \begin{subfigure}{\linewidth}
    \centering
        \begin{minipage}{\linewidth}
        \includegraphics[width=.195\linewidth]{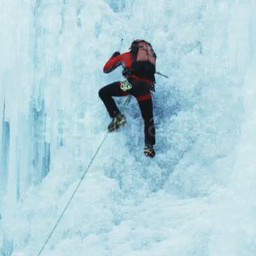}
        \hspace{-0.4em}
        \includegraphics[width=.195\linewidth]{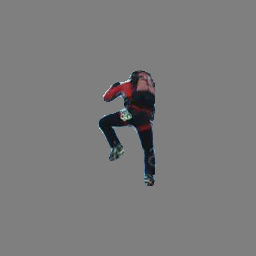}
        \hspace{-0.4em}
        \includegraphics[width=.195\linewidth]{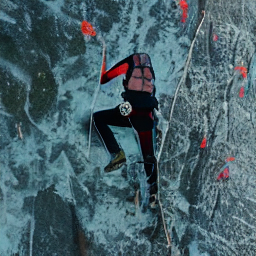}
        \hspace{-0.4em}
        \includegraphics[width=.195\linewidth]{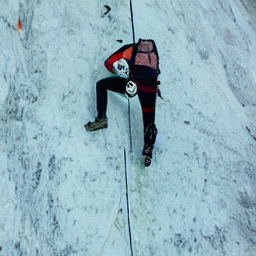}
        \hspace{-0.4em}
        \includegraphics[width=.195\linewidth]{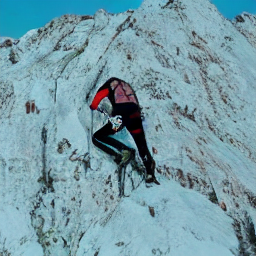}
        \end{minipage}
    \end{subfigure}
    \begin{subfigure}{\linewidth}
    \centering
        \begin{minipage}{\linewidth}
        \includegraphics[width=.195\linewidth]{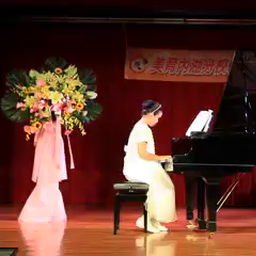}
        \hspace{-0.4em}
        \includegraphics[width=.195\linewidth]{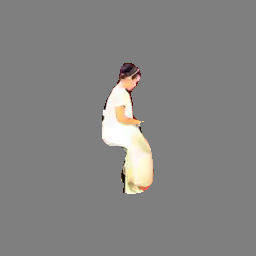}
        \hspace{-0.4em}
        \includegraphics[width=.195\linewidth]{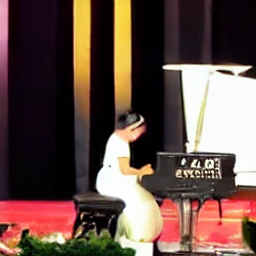}
        \hspace{-0.4em}
        \includegraphics[width=.195\linewidth]{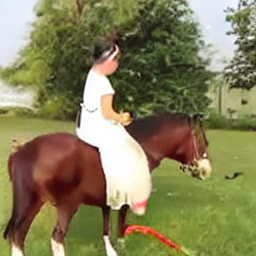}
        \hspace{-0.4em}
        \includegraphics[width=.195\linewidth]{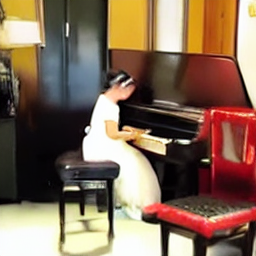}
        \end{minipage}
    \end{subfigure}
    \begin{subfigure}{\linewidth}
    \centering
        \begin{minipage}{\linewidth}
        \includegraphics[width=.195\linewidth]{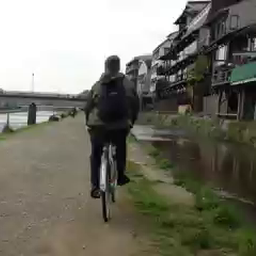}
        \hspace{-0.4em}
        \includegraphics[width=.195\linewidth]{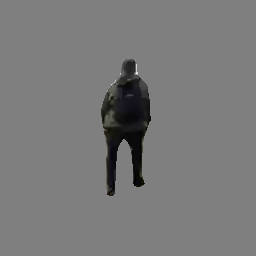}
        \hspace{-0.4em}
        \includegraphics[width=.195\linewidth]{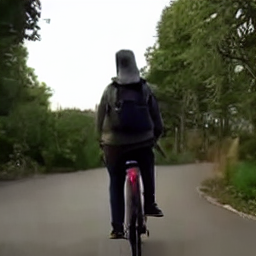}
        \hspace{-0.4em}
        \includegraphics[width=.195\linewidth]{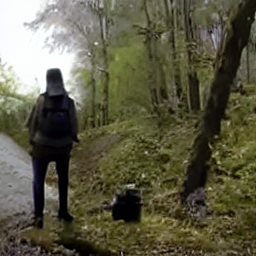}
        \hspace{-0.4em}
        \includegraphics[width=.195\linewidth]{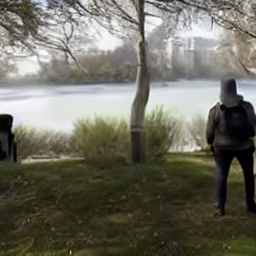}
        \end{minipage}
    \end{subfigure}
    \begin{subfigure}{\linewidth}
    \centering
        \begin{minipage}{\linewidth}
        \includegraphics[width=.195\linewidth]{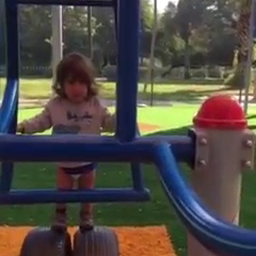}
        \hspace{-0.4em}
        \includegraphics[width=.195\linewidth]{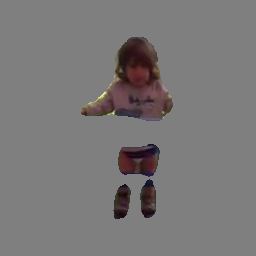}
        \hspace{-0.4em}
        \includegraphics[width=.195\linewidth]{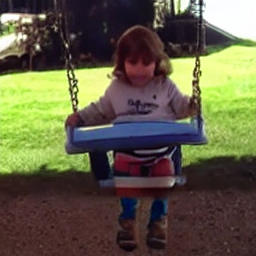}
        \hspace{-0.4em}
        \includegraphics[width=.195\linewidth]{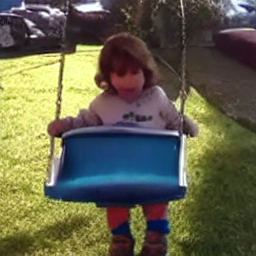}
        \hspace{-0.4em}
        \includegraphics[width=.195\linewidth]{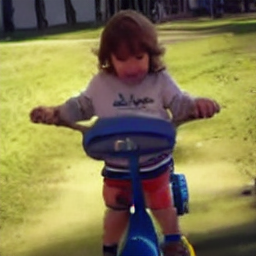}
        \end{minipage}
    \end{subfigure}

\caption{
\textbf{Qualitative results of unconstrained scene hallucination.} Similar to Fig.~\ref{fig:scene-hall-center} but the person here is not constrained on location and pose, hence they can change according to the hallucinated scene. As a result, we are able to generate our reference person in diverse poses while hallucinating different scenes.
}
\vspace{-1.4em}
\label{fig:scene-hall-freeform}
\end{figure}

Quantitative results are shown in Tab.~\ref{tab:ablations}. We observe that image-only models (with or without augmentations) always underperform models trained on video data. This shows that videos provide richer training signal of the same person in different poses which cannot be replicated by simple augmentations. Augmentations, however, do help improve our results. CLIP ViT-L/14 features perform better than the VAE features passed through concatenation. We also note that using a larger 860M UNet and initializing with Stable Diffusion checkpoints help with our model performance.

We present qualitative results for our best-performing model in Fig.~\ref{fig:res-cond-fig}. In the top four rows, we show how our model can infer candidate poses given scene context and flexibly re-pose the same reference person into various different scenes. In the bottom four rows, we also show how different people can coherently be inserted into the same scene. The generated images show the complex human-scene composition learned by our model. Our model also harmonizes the insertion by accounting for lighting and shadows.

\xhdr{Effect of CFG.} We present the metric trend with varying CFG~\cite{ho2022classifier} guidance scales in Fig.~\ref{fig:cfg-person-conditioning}. In line with observations from text-to-image models~\cite{saharia2022photorealistic, rombach2022high}, our FID and PCKh both initially improve with CFG. At high values, the image quality (FID) suffers. We perform CFG by dropping both the masked scene image and reference person to learn a true unconditional distribution. We observed that dropping only the reference person was detrimental to our model performance.

\vspace{-0.2em}
\subsection{Person Hallucination}
\label{sec:exp-human-hall}
\vspace{-0.2em}

We evaluate the person hallucination task by dropping the person conditioning and compare with baselines Stable Diffusion~\cite{stablediffusion} and DALL-E 2~\cite{ramesh2022hierarchical}. We evaluate our model by passing an empty conditioning person. We evaluate quantitatively with Stable Diffusion (SD) with the following prompt: ``natural coherent image of a person in a scene''. For qualitative evaluation, we generate SD and DALL-E 2 results with the same prompt.

\begin{table}
\caption{Comparison of metrics with Stable Diffusion for person and scene hallucination. For Stable Diffusion. we use the following prompt: "a natural coherent image of a person in a scene.".
}

\small
\centering\vspace{5pt}
\begin{tabular}{lccc}

\toprule

\multirow{2}{*}{Method} & \multicolumn{2}{c}{Person hall.} & Scene hall.\\
\cmidrule(lr){2-3}\cmidrule(lr){4-4}
& FID $\downarrow$ & PCKh $\uparrow$  & FID $\downarrow$ \\
\midrule

Stable Diffusion & 19.651 & 0.023 & 44.687 \\
Ours &  8.390 & 23.213 & 20.268  \\

\bottomrule
\vspace{-2.0em}
\end{tabular}
\label{tab:scene-hall}
\end{table}

We present qualitative results in Fig.~\ref{fig:res-person-hall} where our model can successfully hallucinate diverse people given a masked scene image. The hallucinated people have poses consistent with the input scene affordances. We also present quantitative results in Tab.~\ref{tab:scene-hall}. While Stable Diffusion does produce high-quality results in some cases, it sometimes fails catastrophically such as hallucinating a person in incorrect poses, or completely ignoring the text conditioning. This is expected as Stable Diffusion's inpainting is trained to inpaint random crops with generic captions rather than inpainting with a consistent human, which is our objective. 

We present qualitative baseline comparisons in Fig.~\ref{fig:person-hall-baselines}, we observe that baseline models sometimes ignore the scene context while our model does better at hallucinating humans consistent with the scene.

\xhdr{Effect of CFG.} The metric trend with varying CFG scales for person hallucination follows closely with the person conditioning trend, as shown in Fig.~\ref{fig:cfg-person-hallucination}. Both FID and PCKh initially improve, after which FID worsens.

\vspace{-0.2em}
\subsection{Scene Hallucination}
\label{sec:exp-scene-hall}
\vspace{-0.2em}

We evaluate two kinds of scene hallucination tasks. \textbf{Constrained:} For the constrained setup, we pass the reference person as the scene image. The model then retains the location and pose of the person and hallucinates a consistent scene around the person. \textbf{Unconstrained:} For the unconstrained setup, we pass an empty scene conditioning. Given a reference person, the model then simultaneously hallucinates a scene and places the person in the right location and pose.  We evaluate the constrained setup quantitatively with SD with the same prompt as before. We also present qualitative samples from SD and DALL-E 2.

We present qualitative results of the constrained case in Fig.~\ref{fig:scene-hall-center} and unconstrained case in Fig.~\ref{fig:scene-hall-freeform}. Quantitative comparisons are in Tab.~\ref{tab:scene-hall}. As hallucinating scenes is a harder task with large portions of the image to be synthesized, FID scores are generally higher with our model performing better. Some qualitative baseline comparisons are presented in Fig.~\ref{fig:scene-hall-baselines}. Compared to the baselines, our model synthesizes more realistic scenes while maintaining coherence with the input reference person.

\label{sec:exp-failure}

\vspace{-0.2em}
\section{Conclusion}
\label{sec:conclusion}
\vspace{-0.2em}
In this work, we propose a novel task of affordance-aware human insertion into scenes and we solve it by learning a conditional diffusion model in a self-supervised way using video data. We show various qualitative results to demonstrate the effectiveness of our approach. We also perform detailed ablation studies to analyze the impacts of various design choices. We hope this work will inspire other researchers to pursue this new research direction.

\vspace{-0.6em}

{\small \paragraph{Acknowledgement.} We are grateful to Fabian Caba Heilbron, Tobias Hinz, Kushal Kafle, Nupur Kumari, Yijun Li and Markus Woodson for insightful discussions regarding data and training pipeline. This work was partly done by Sumith Kulal during an internship at Adobe Research. Additional funding provided by ONR MURI and NSF GRFP.}

{\small
\bibliography{egbib}
\bibliographystyle{ieee_fullname}
}

\end{document}


\title{Putting People in Their Place: Affordance-Aware Human Insertion into Scenes \normalfont{Supplementary Materials}\vspace{-0.5em}}
\author{
Sumith Kulal$^{1}$\\
\vspace{-1em}
\and
Tim Brooks$^{2}$
\and
Alex Aiken$^{1}$
\and
Jiajun Wu$^{1}$
\and
Jimei Yang$^{3}$
\and
Jingwan Lu$^{3}$
\and
Alexei A.\ Efros$^{2}$
\and
Krishna Kumar Singh$^{3}$
}

\twocolumn[{
\maketitle
\vspace{-2.8em}
\begin{center}
{\large
$^{1}$Stanford University\hspace{0.8em}$^{2}$UC Berkeley\hspace{0.8em}$^{3}$Adobe Research
}
\end{center}
\vspace{1em}
}]

\begin{figure*}

\centering
    \begin{subfigure}{\linewidth}
    \centering
        \begin{minipage}{\linewidth}
        \includegraphics[width=.105\linewidth]{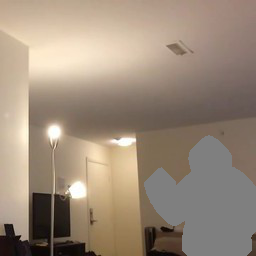}
        \includegraphics[width=.105\linewidth]{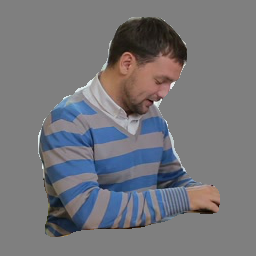}
        \hspace{-0.4em}
        \includegraphics[width=.105\linewidth]{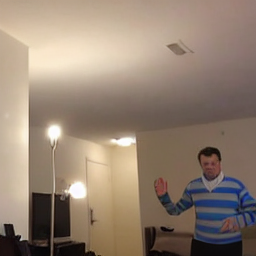}
        \includegraphics[width=.105\linewidth]{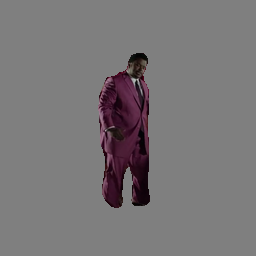}
        \hspace{-0.4em}
        \includegraphics[width=.105\linewidth]{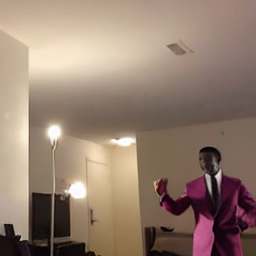}
        \includegraphics[width=.105\linewidth]{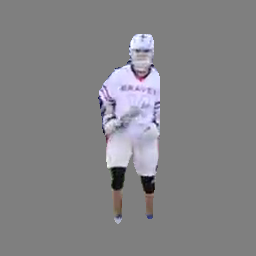}
        \hspace{-0.4em}
        \includegraphics[width=.105\linewidth]{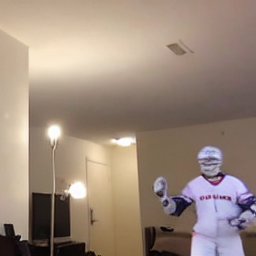}
        \includegraphics[width=.105\linewidth]{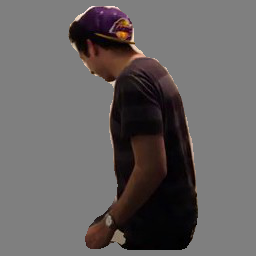}
        \hspace{-0.4em}
        \includegraphics[width=.105\linewidth]{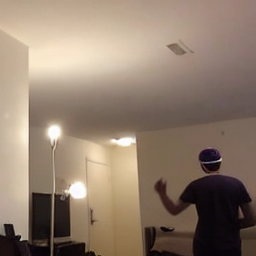}
        \end{minipage}
    \end{subfigure}
    \begin{subfigure}{\linewidth}
    \centering
        \begin{minipage}{\linewidth}
        \includegraphics[width=.105\linewidth]{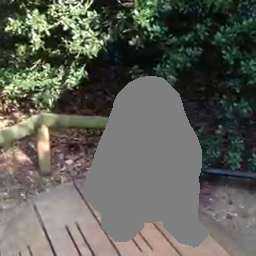}
        \includegraphics[width=.105\linewidth]{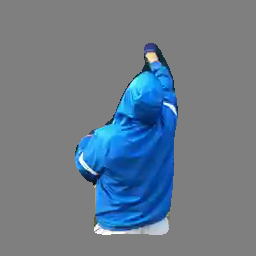}
        \hspace{-0.4em}
        \includegraphics[width=.105\linewidth]{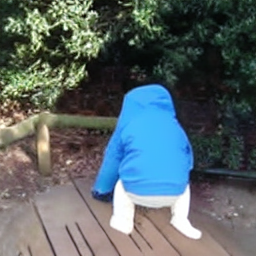}
        \includegraphics[width=.105\linewidth]{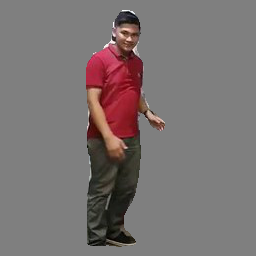}
        \hspace{-0.4em}
        \includegraphics[width=.105\linewidth]{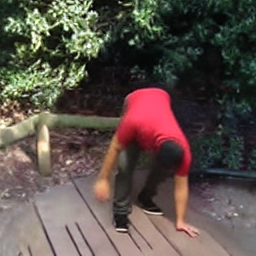}
        \includegraphics[width=.105\linewidth]{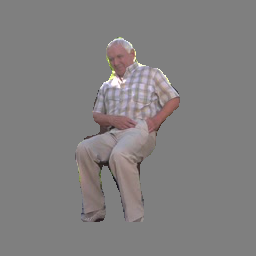}
        \hspace{-0.4em}
        \includegraphics[width=.105\linewidth]{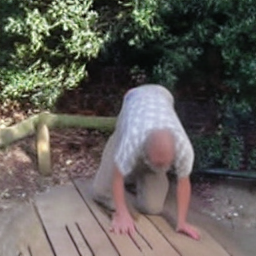}
        \includegraphics[width=.105\linewidth]{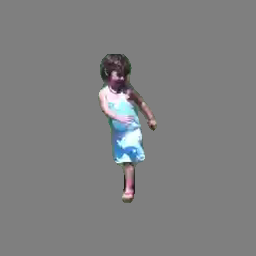}
        \hspace{-0.4em}
        \includegraphics[width=.105\linewidth]{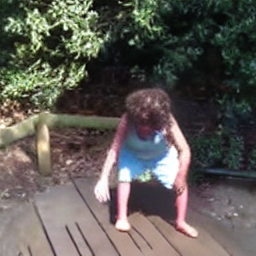}
        \end{minipage}
    \end{subfigure}
    \begin{subfigure}{\linewidth}
    \centering
        \begin{minipage}{\linewidth}
        \includegraphics[width=.105\linewidth]{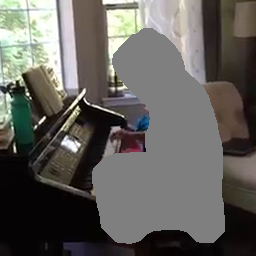}
        \includegraphics[width=.105\linewidth]{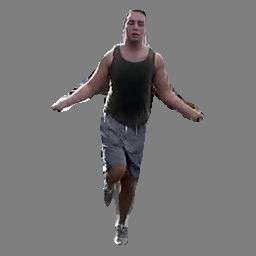}
        \hspace{-0.4em}
        \includegraphics[width=.105\linewidth]{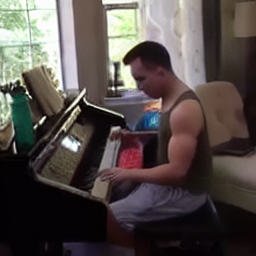}
        \includegraphics[width=.105\linewidth]{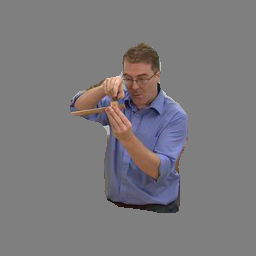}
        \hspace{-0.4em}
        \includegraphics[width=.105\linewidth]{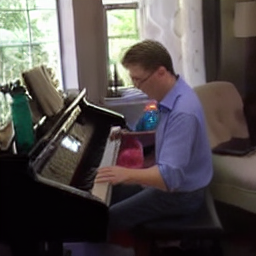}
        \includegraphics[width=.105\linewidth]{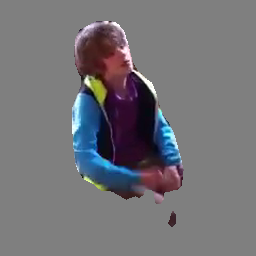}
        \hspace{-0.4em}
        \includegraphics[width=.105\linewidth]{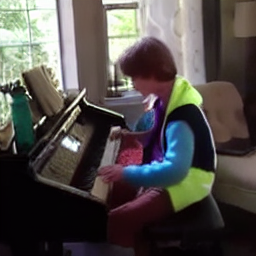}
        \includegraphics[width=.105\linewidth]{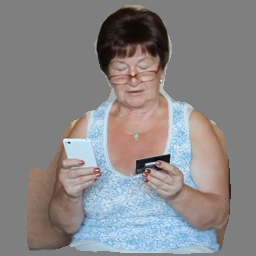}
        \hspace{-0.4em}
        \includegraphics[width=.105\linewidth]{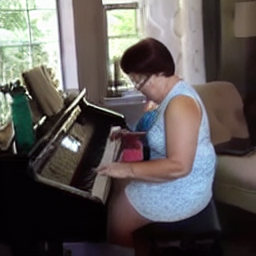}
        \end{minipage}
    \end{subfigure}
    \begin{subfigure}{\linewidth}
    \centering
        \begin{minipage}{\linewidth}
        \includegraphics[width=.105\linewidth]{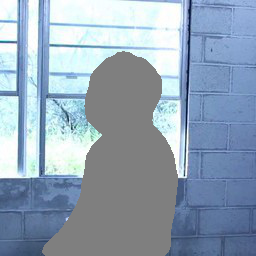}
        \includegraphics[width=.105\linewidth]{fig/cond_seg_gen/person_01009.png}
        \hspace{-0.4em}
        \includegraphics[width=.105\linewidth]{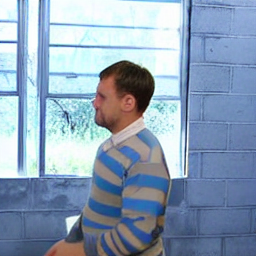}
        \includegraphics[width=.105\linewidth]{fig/cond_seg_gen/person_01031.png}
        \hspace{-0.4em}
        \includegraphics[width=.105\linewidth]{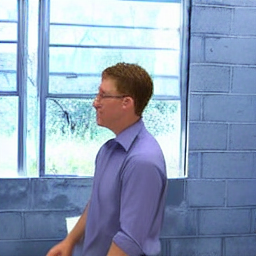}
        \includegraphics[width=.105\linewidth]{fig/cond_seg_gen/person_01488.png}
        \hspace{-0.4em}
        \includegraphics[width=.105\linewidth]{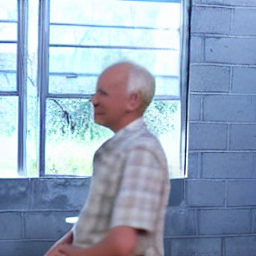}
        \includegraphics[width=.105\linewidth]{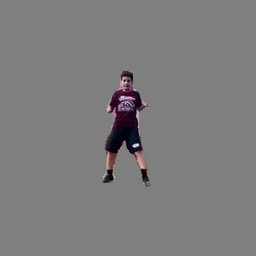}
        \hspace{-0.4em}
        \includegraphics[width=.105\linewidth]{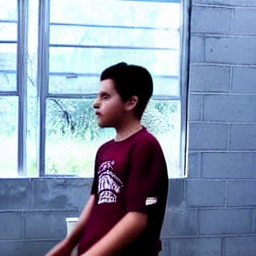}

        \end{minipage}
    \end{subfigure}

\caption{
\textbf{Qualitative results of conditional generation on segmentation masks.} In the first column, we show the scene image with a segmentation mask. After that, we show the results of inserting four different people in the scene image. For each result, we first show the person to be inserted followed by our insertion result. We can see that our inserted person follows the segmentation mask while being coherent with the scene.  
}
\label{fig:cond-seg-gen-fig}
\end{figure*}
We discuss implementation details in Sec.~\ref{sec:implementation}. We present results on general segmentation masks instead of bounding boxes in Sec.~\ref{sec:seg-masks}, partial body completion in Sec.~\ref{sec:partial-body}, and cloth swapping in Sec.~\ref{sec:cloth-swap}. We highlight the diversity of poses predicted by our model in Sec.~\ref{sec:diversity}. We conclude with a discussion of failure cases in Sec.~\ref{sec:failure} and societal impact in Sec.~\ref{sec:societal}.

\section{Implementation details}
\label{sec:implementation}

We use the Stable Diffusion~\cite{stablediffusion} architecture as our backbone and leverage their pre-trained weights as our network initialization. During the forward process, we use a linear noise schedule for $1000$ noising steps in the interval $[0.00085, 0.0120]$. During the reverse process, at inference time, we use DDIM sampler~\cite{song2021ddim} for 200 steps.

As in the Stable Diffusion architecure, our model uses the first stage VAE to encode $256\times256\times3$ image into $32\times32\times4$ latents. The denoising UNet has a convolution encoder that transforms the $9$ channel input with noisy image, masked image, and mask to a $320$ channel embedding. The multiplication factors of our UNet are $[1, 2, 4, 4]$. In the input blocks, the height and width get scaled down by the factor and the channel dimension gets scaled up by the same factor. The output blocks do the opposite. Self-attention and cross-attention with conditioning are present at layers $8\times8$, $16\times16$, and $32\times32$ with $8$ attention heads.

\subsection{Masking details}
The configuration of the masking strategy used is:

\begin{itemize}[noitemsep,topsep=0pt,parsep=0pt,partopsep=0pt,leftmargin=2em]
\item \textbf{Bounding box}: 30\% of the time, we use randomly dilated (0 upto 20 pixels) person bounding box. 
\item \textbf{Larger boxes}: 20\% of the time, we randomly sample a larger bounding box (5-20\% larger in area) that contains the person bounding box. 
\item \textbf{Random boxes}: 15\% of the time, we randomly sample a smaller bounding box (25-75\% area) within the person bounding box. 
\item \textbf{Person segmentation}: 15\% of the time, we use randomly dilated (0 upto 20 pixels) person segmentation masks.
\item \textbf{Random scribbles}: 20\% of the time, we use randomly generated scribble and brush masks as done in prior inpainting works~\cite{zhao2021large, yu2019free}.
\end{itemize}

\subsection{Augmentation details}

We apply augmentation on the reference person alone. Given a reference person, we first apply color augmentations. We then mask and center the person. We then apply geometric augmentations. Our augmentation pipeline closely follows StyleGAN-ADA~\cite{karras2020training}.

For color augmentations, we perform brightness, contrast, saturation, image-space filtering and additive noise with probabilities of $0.2$, $0.2$, $0.2$, $0.1$ and $0.1$ respectively. For geometric augmentations, we perform isotropic scaling, rotation, anistropic scaling and cutout with probabilities $0.4$, $0.4$, $0.2$ and $0.2$.

\section{Segmentation mask results}
\label{sec:seg-masks}
We present results on person segmentation masks as holes in Fig.~\ref{fig:cond-seg-gen-fig} to demonstrate support for arbitrary shaped masks in addition to rectangular bounding boxes.

\begin{figure*}

\centering

    \begin{subfigure}{\linewidth}
    \centering
        \begin{minipage}{\linewidth}
        \includegraphics[width=.105\linewidth]{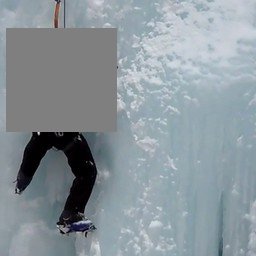}
        \includegraphics[width=.105\linewidth]{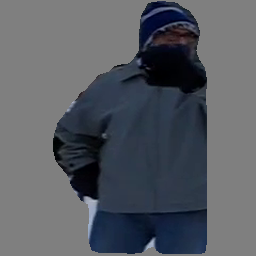}
        \hspace{-0.4em}
        \includegraphics[width=.105\linewidth]{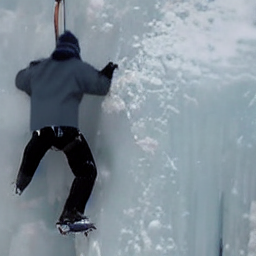}
        \includegraphics[width=.105\linewidth]{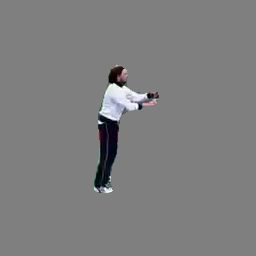}
        \hspace{-0.4em}
        \includegraphics[width=.105\linewidth]{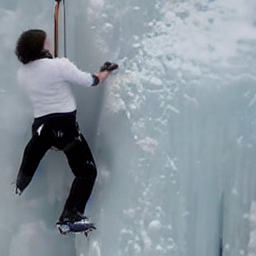}
        \includegraphics[width=.105\linewidth]{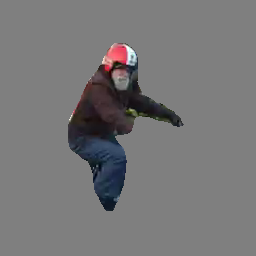}
        \hspace{-0.4em}
        \includegraphics[width=.105\linewidth]{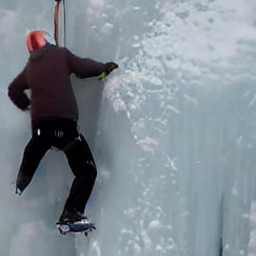}
        \includegraphics[width=.105\linewidth]{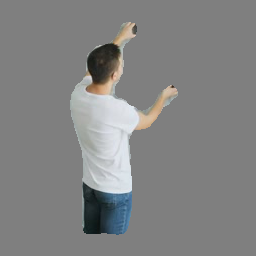}
        \hspace{-0.4em}
        \includegraphics[width=.105\linewidth]{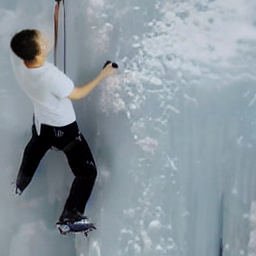}
        \end{minipage}
    \end{subfigure}
    \begin{subfigure}{\linewidth}
    \centering
        \begin{minipage}{\linewidth}
        \includegraphics[width=.105\linewidth]{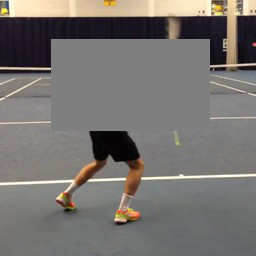}
        \includegraphics[width=.105\linewidth]{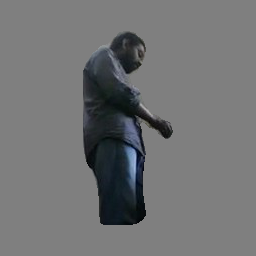}
        \hspace{-0.4em}
        \includegraphics[width=.105\linewidth]{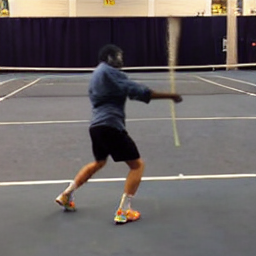}
        \includegraphics[width=.105\linewidth]{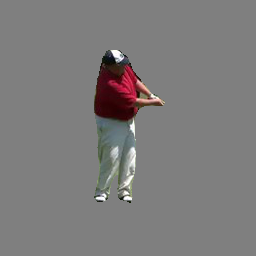}
        \hspace{-0.4em}
        \includegraphics[width=.105\linewidth]{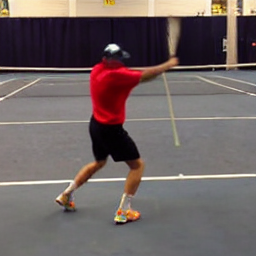}
        \includegraphics[width=.105\linewidth]{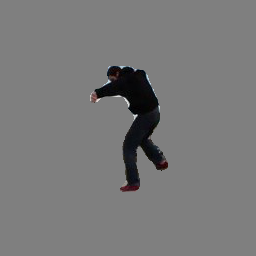}
        \hspace{-0.4em}
        \includegraphics[width=.105\linewidth]{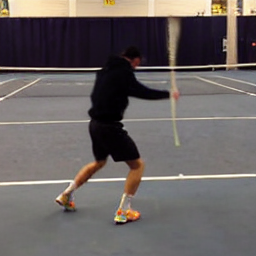}
        \includegraphics[width=.105\linewidth]{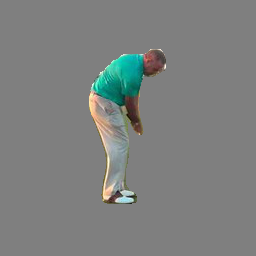}
        \hspace{-0.4em}
        \includegraphics[width=.105\linewidth]{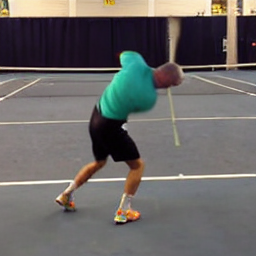}
        \end{minipage}
    \end{subfigure}
    \begin{subfigure}{\linewidth}
    \centering
        \begin{minipage}{\linewidth}
        \includegraphics[width=.105\linewidth]{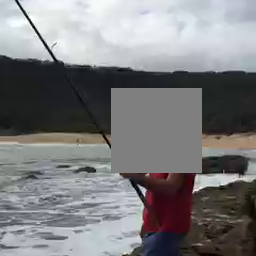}
        \includegraphics[width=.105\linewidth]{fig/cond_halves_gen/person_01002.png}
        \hspace{-0.4em}
        \includegraphics[width=.105\linewidth]{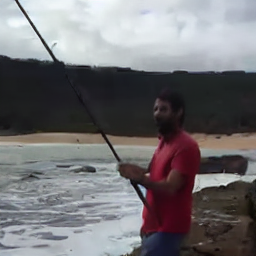}
        \includegraphics[width=.105\linewidth]{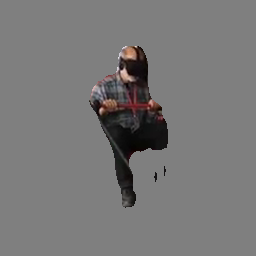}
        \hspace{-0.4em}
        \includegraphics[width=.105\linewidth]{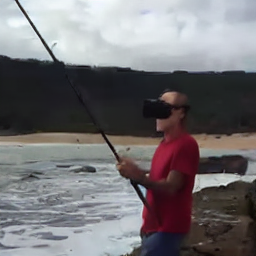}
        \includegraphics[width=.105\linewidth]{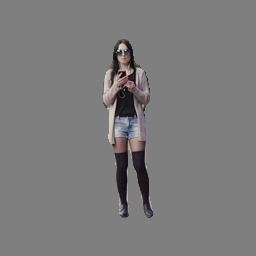}
        \hspace{-0.4em}
        \includegraphics[width=.105\linewidth]{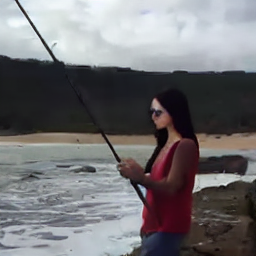}
        \includegraphics[width=.105\linewidth]{fig/cond_halves_gen/person_01827.png}
        \hspace{-0.4em}
        \includegraphics[width=.105\linewidth]{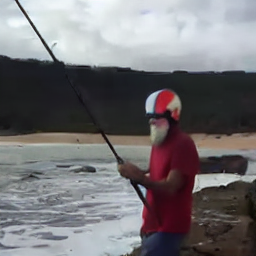}
        \end{minipage}
    \end{subfigure}
    \begin{subfigure}{\linewidth}
    \centering
        \begin{minipage}{\linewidth}
        \includegraphics[width=.105\linewidth]{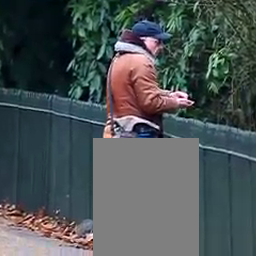}
        \includegraphics[width=.105\linewidth]{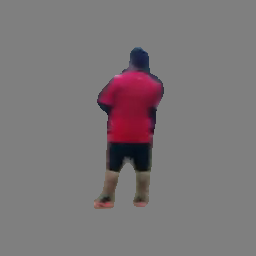}
        \hspace{-0.4em}
        \includegraphics[width=.105\linewidth]{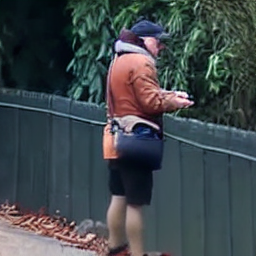}
        \includegraphics[width=.105\linewidth]{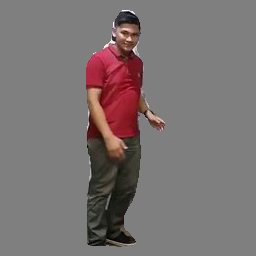}
        \hspace{-0.4em}
        \includegraphics[width=.105\linewidth]{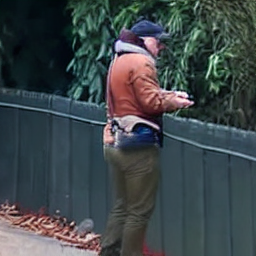}
        \includegraphics[width=.105\linewidth]{fig/cond_halves_gen/person_01366.png}
        \hspace{-0.4em}
        \includegraphics[width=.105\linewidth]{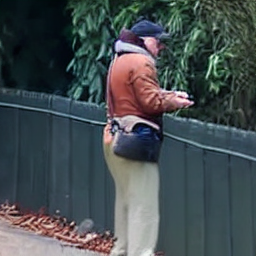}
        \includegraphics[width=.105\linewidth]{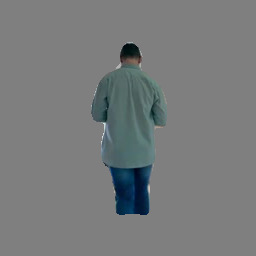}
        \hspace{-0.4em}
        \includegraphics[width=.105\linewidth]{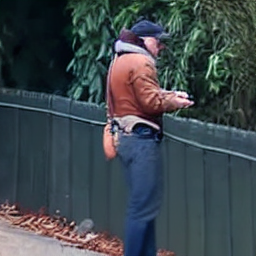}
        \end{minipage}
    \end{subfigure}

\caption{
\textbf{Qualitative results of partial body completion.}  In the first column, we show the scene image with only the partial body masked. After that, we show the results of inserting four different people in the scene image. For each result, we first show the person to be inserted followed by our insertion result. We can see that our inserted person is consistent with the visible partial body in terms of the pose while retaining its original appearance.
}
\label{fig:cond-halves-gen-fig}
\end{figure*}
\begin{figure*}

\centering
    \begin{subfigure}{\linewidth}
    \centering
        \begin{minipage}{\linewidth}
        \includegraphics[width=.105\linewidth]{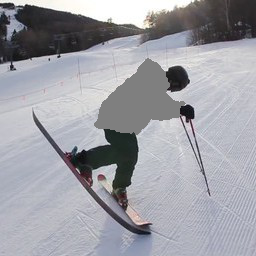}
        \includegraphics[width=.105\linewidth]{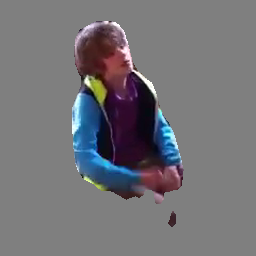}
        \hspace{-0.4em}
        \includegraphics[width=.105\linewidth]{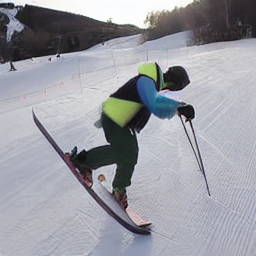}
        \includegraphics[width=.105\linewidth]{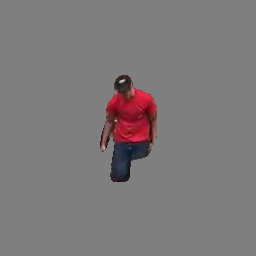}
        \hspace{-0.4em}
        \includegraphics[width=.105\linewidth]{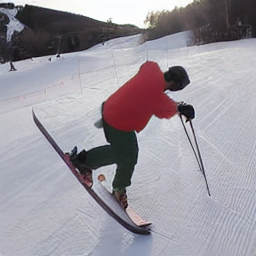}
        \includegraphics[width=.105\linewidth]{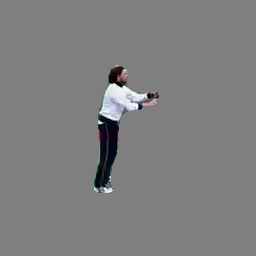}
        \hspace{-0.4em}
        \includegraphics[width=.105\linewidth]{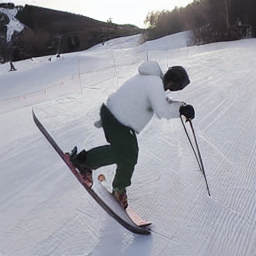}
        \includegraphics[width=.105\linewidth]{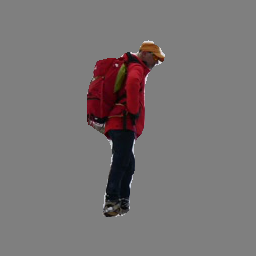}
        \hspace{-0.4em}
        \includegraphics[width=.105\linewidth]{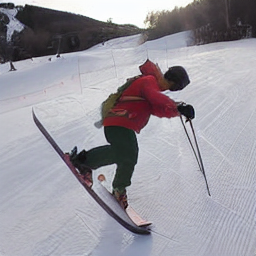}
        \end{minipage}
    \end{subfigure}
    \begin{subfigure}{\linewidth}
    \centering
        \begin{minipage}{\linewidth}
        \includegraphics[width=.105\linewidth]{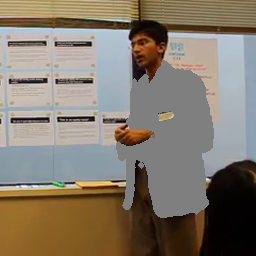}
        \includegraphics[width=.105\linewidth]{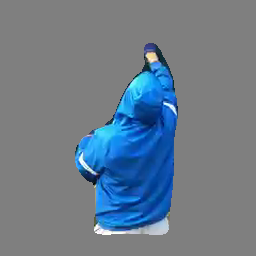}
        \hspace{-0.4em}
        \includegraphics[width=.105\linewidth]{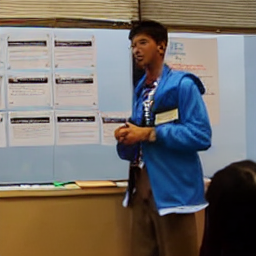}
        \includegraphics[width=.105\linewidth]{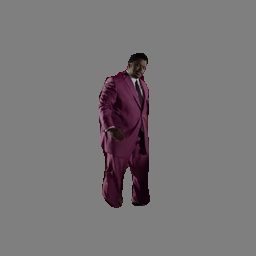}
        \hspace{-0.4em}
        \includegraphics[width=.105\linewidth]{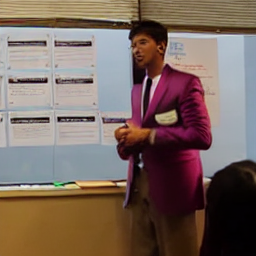}
        \includegraphics[width=.105\linewidth]{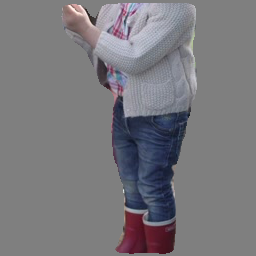}
        \hspace{-0.4em}
        \includegraphics[width=.105\linewidth]{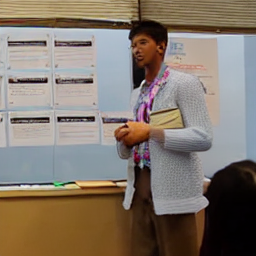}
        \includegraphics[width=.105\linewidth]{fig/cond_shirts_gen/person_01547.png}
        \hspace{-0.4em}
        \includegraphics[width=.105\linewidth]{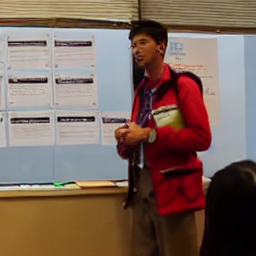}
        \end{minipage}
    \end{subfigure}
    \begin{subfigure}{\linewidth}
    \centering
        \begin{minipage}{\linewidth}
        \includegraphics[width=.105\linewidth]{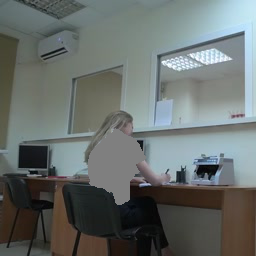}
        \includegraphics[width=.105\linewidth]{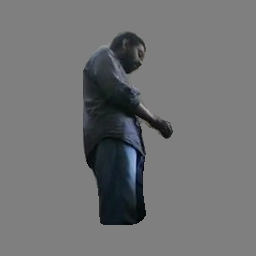}
        \hspace{-0.4em}
        \includegraphics[width=.105\linewidth]{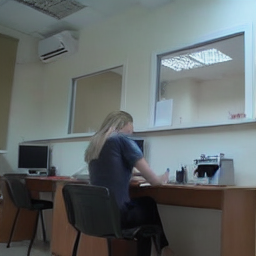}
        \includegraphics[width=.105\linewidth]{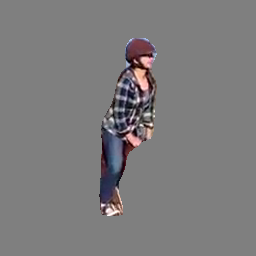}
        \hspace{-0.4em}
        \includegraphics[width=.105\linewidth]{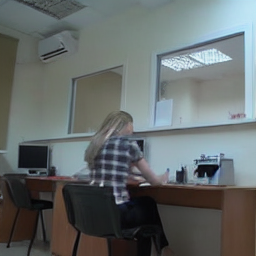}
        \includegraphics[width=.105\linewidth]{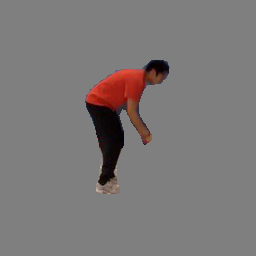}
        \hspace{-0.4em}
        \includegraphics[width=.105\linewidth]{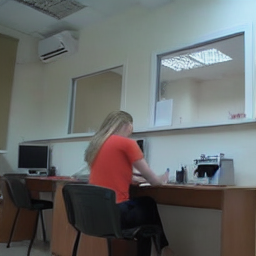}
        \includegraphics[width=.105\linewidth]{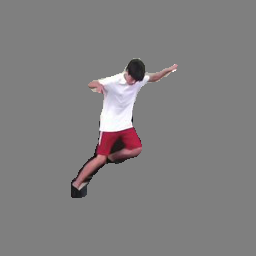}
        \hspace{-0.4em}
        \includegraphics[width=.105\linewidth]{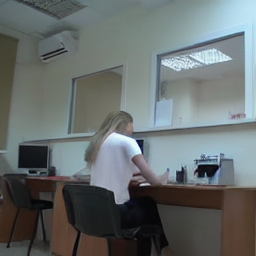}
        \end{minipage}
    \end{subfigure}
    \begin{subfigure}{\linewidth}
    \centering
        \begin{minipage}{\linewidth}
        \includegraphics[width=.105\linewidth]{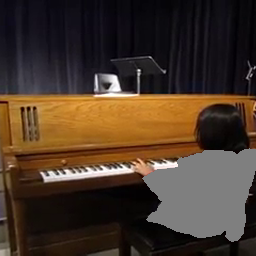}
        \includegraphics[width=.105\linewidth]{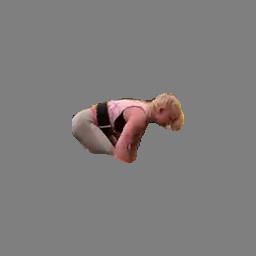}
        \hspace{-0.4em}
        \includegraphics[width=.105\linewidth]{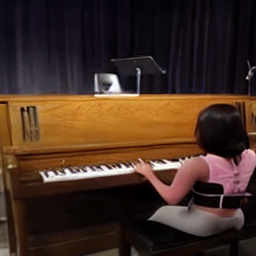}
        \includegraphics[width=.105\linewidth]{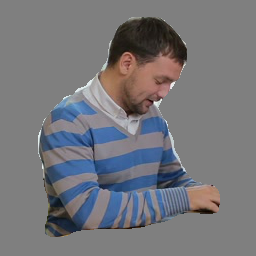}
        \hspace{-0.4em}
        \includegraphics[width=.105\linewidth]{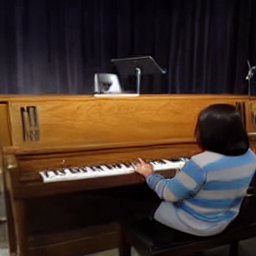}
        \includegraphics[width=.105\linewidth]{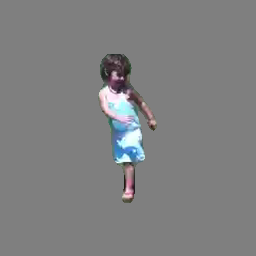}
        \hspace{-0.4em}
        \includegraphics[width=.105\linewidth]{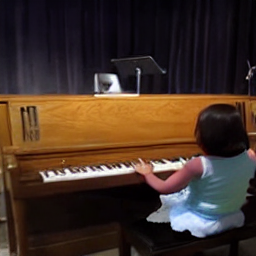}
        \includegraphics[width=.105\linewidth]{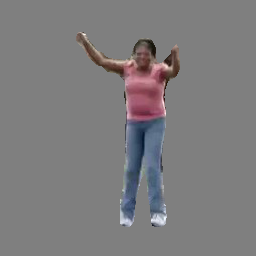}
        \hspace{-0.4em}
        \includegraphics[width=.105\linewidth]{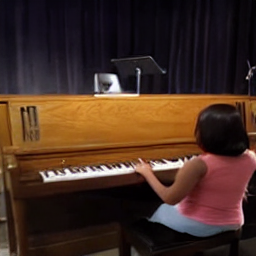}
        \end{minipage}
    \end{subfigure}

\caption{
\textbf{Qualitative results of cloth swapping.} 
In the first column, we show the scene image with only the upper body cloth masked. After that, we show the results of inserting four different people in the scene image. For each result, we first show the person to be inserted followed by our insertion result. We can see that our generated result is successfully able to borrow the upper body cloth from the person to be inserted. Also, these cloth swaps were quite challenging due to differences in the pose, viewpoint, and scale between the person in the scene image and the person to be inserted. 
}
\label{fig:cond-shirts-gen-fig}
\end{figure*}
\begin{figure}
\centering

    \begin{subfigure}{\linewidth}
    \centering
        \begin{minipage}{\linewidth}
        \includegraphics[width=.195\linewidth]{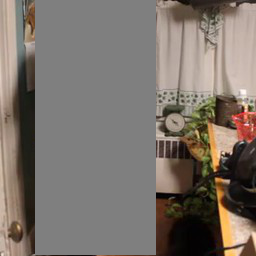}
        \hspace{-0.4em}
        \includegraphics[width=.195\linewidth]{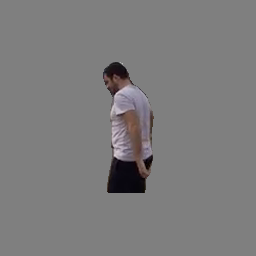}
        \hspace{-0.4em}
            \includegraphics[width=.195\linewidth]{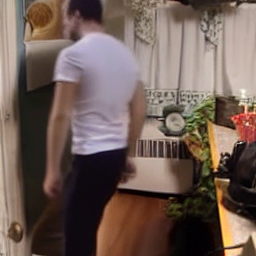}
        \hspace{-0.4em}
            \includegraphics[width=.195\linewidth]{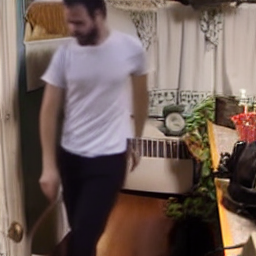}
        \hspace{-0.4em}
            \includegraphics[width=.195\linewidth]{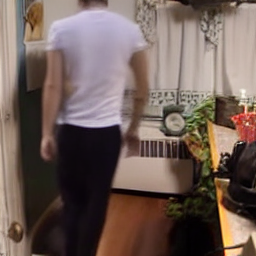}
        \end{minipage}
    \end{subfigure}
    \begin{subfigure}{\linewidth}
    \centering
        \begin{minipage}{\linewidth}
        \includegraphics[width=.195\linewidth]{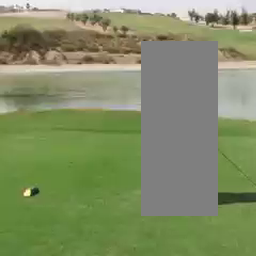}
        \hspace{-0.4em}
        \includegraphics[width=.195\linewidth]{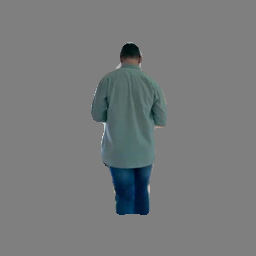}
        \hspace{-0.4em}
            \includegraphics[width=.195\linewidth]{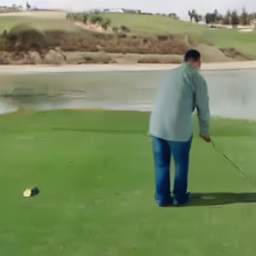}
        \hspace{-0.4em}
            \includegraphics[width=.195\linewidth]{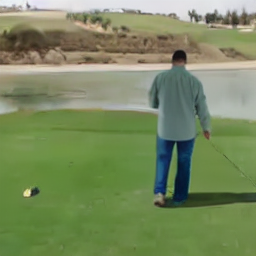}
        \hspace{-0.4em}
            \includegraphics[width=.195\linewidth]{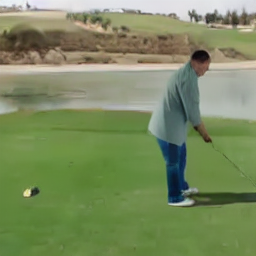}
        \end{minipage}
    \end{subfigure}
    \begin{subfigure}{\linewidth}
    \centering
        \begin{minipage}{\linewidth}
        \includegraphics[width=.195\linewidth]{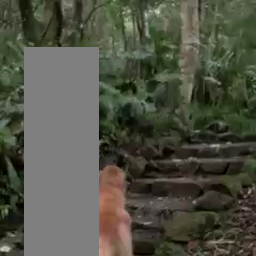}
        \hspace{-0.4em}
        \includegraphics[width=.195\linewidth]{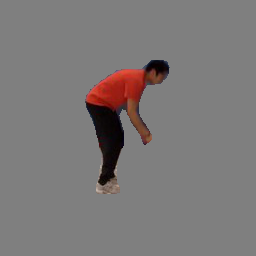}
        \hspace{-0.4em}
            \includegraphics[width=.195\linewidth]{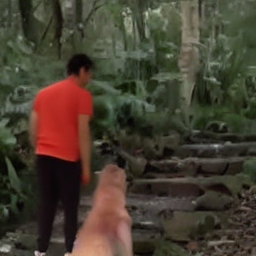}
        \hspace{-0.4em}
            \includegraphics[width=.195\linewidth]{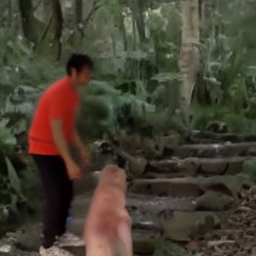}
        \hspace{-0.4em}
            \includegraphics[width=.195\linewidth]{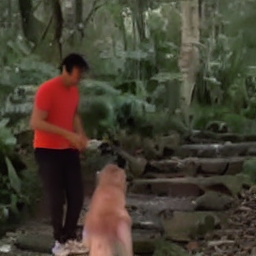}
        \end{minipage}
    \end{subfigure}
    \begin{subfigure}{\linewidth}
    \centering
        \begin{minipage}{\linewidth}
        \includegraphics[width=.195\linewidth]{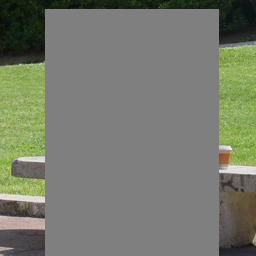}
        \hspace{-0.4em}
        \includegraphics[width=.195\linewidth]{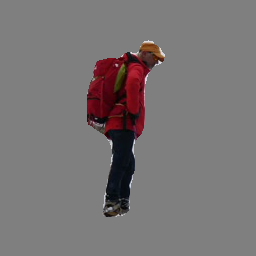}
        \hspace{-0.4em}
            \includegraphics[width=.195\linewidth]{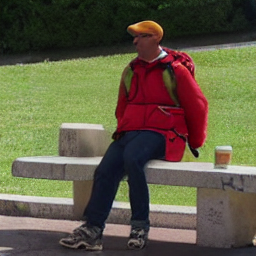}
        \hspace{-0.4em}
            \includegraphics[width=.195\linewidth]{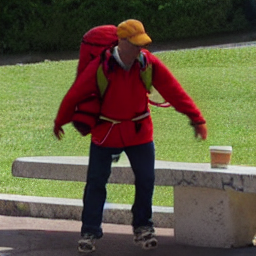}
        \hspace{-0.4em}
            \includegraphics[width=.195\linewidth]{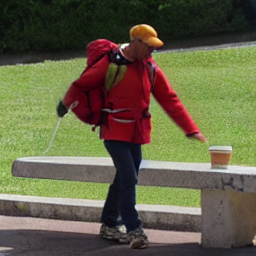}
        \end{minipage}
    \end{subfigure}

\vspace{-0.5em}
\caption{
\textbf{Diverse generation for same masked scene image and reference person to be inserted.}  In the first column, we show the masked scene image, followed by the reference person to be inserted. After that, we show three different variations of the same person inserted in the scene image. Each variation corresponds to a different noise map during inference. We can see, we are able to compose the same person in the masked region in multiple meaningful ways.
}
\label{fig:cond-diverse-gen-fig}
\vspace{-1.5em}
\end{figure}
\begin{figure}
\centering
    \begin{subfigure}{\linewidth}
    \centering
        \begin{minipage}{\linewidth}
        \includegraphics[width=.195\linewidth]{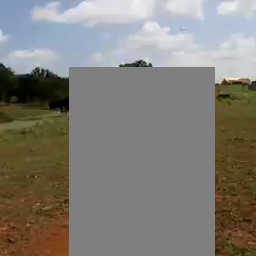}
        \hspace{-0.4em}
        \includegraphics[width=.195\linewidth]{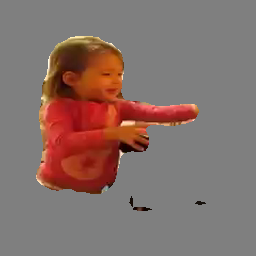}
        \hspace{-0.4em}
        \includegraphics[width=.195\linewidth]{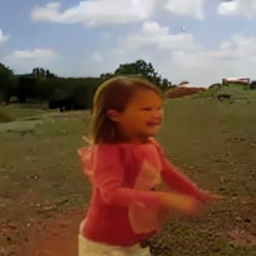}
        \hspace{-0.4em}
        \includegraphics[width=.195\linewidth]{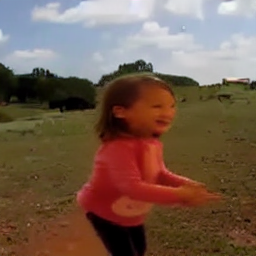}
        \hspace{-0.4em}
        \includegraphics[width=.195\linewidth]{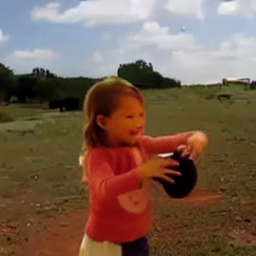}
        \end{minipage}
    \end{subfigure}

    \begin{subfigure}{\linewidth}
    \centering
        \begin{minipage}{\linewidth}
        \includegraphics[width=.195\linewidth]{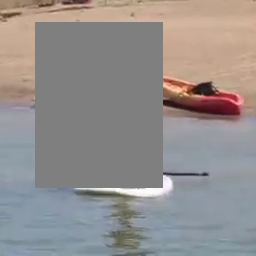}
        \hspace{-0.4em}
        \includegraphics[width=.195\linewidth]{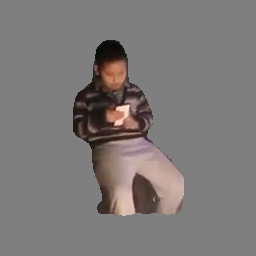}
        \hspace{-0.4em}
        \includegraphics[width=.195\linewidth]{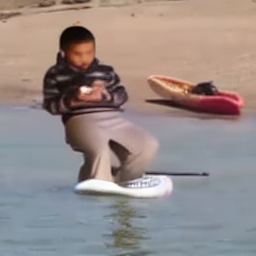}
        \hspace{-0.4em}
        \includegraphics[width=.195\linewidth]{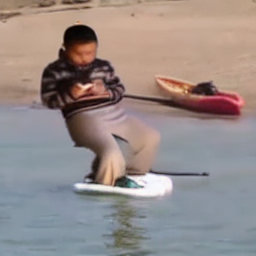}
        \hspace{-0.4em}
        \includegraphics[width=.195\linewidth]{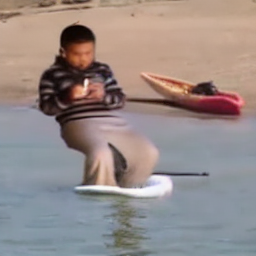}
        \end{minipage}
    \end{subfigure}
    
    \begin{subfigure}{\linewidth}
    \centering
        \begin{minipage}{\linewidth}
        \includegraphics[width=.195\linewidth]{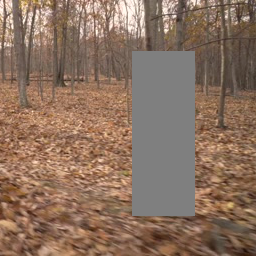}
        \hspace{-0.4em}
        \includegraphics[width=.195\linewidth]{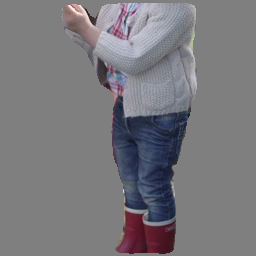}
        \hspace{-0.4em}
        \includegraphics[width=.195\linewidth]{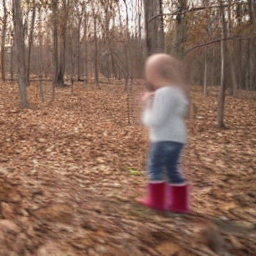}
        \hspace{-0.4em}
        \includegraphics[width=.195\linewidth]{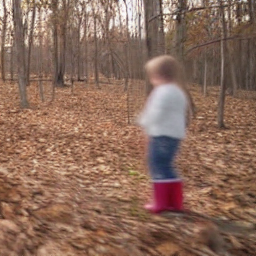}
        \hspace{-0.4em}
        \includegraphics[width=.195\linewidth]{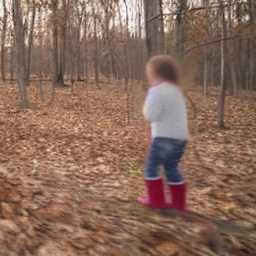}
        \end{minipage}
    \end{subfigure}
    \begin{subfigure}{\linewidth}
    \centering
        \begin{minipage}{\linewidth}
        \includegraphics[width=.195\linewidth]{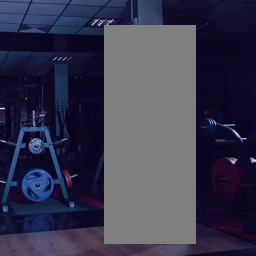}
        \hspace{-0.4em}
        \includegraphics[width=.195\linewidth]{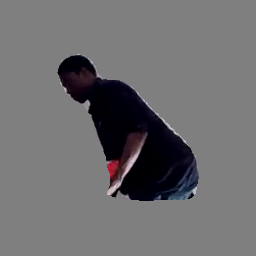}
        \hspace{-0.4em}
        \includegraphics[width=.195\linewidth]{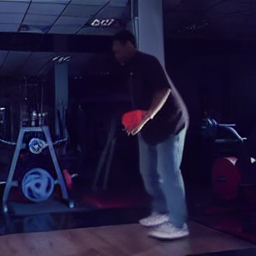}
        \hspace{-0.4em}
        \includegraphics[width=.195\linewidth]{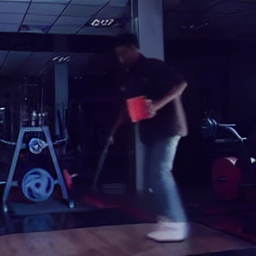}
        \hspace{-0.4em}
        \includegraphics[width=.195\linewidth]{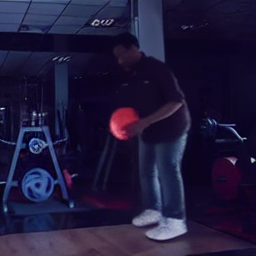}
        \end{minipage}
    \end{subfigure}

\vspace{-0.5em}
\caption{
\textbf{Failure cases.} Our common failure modes are generating bad faces (row 1 \& 2), poor lighning (row 1), extreme pose (row 2), blurry samples (row 3), and generating the object present in the reference person (row 4).
}
\label{fig:cond-failure-gen-fig}
\vspace{-1.5em}
\end{figure}

\section{Partial body completion}
\label{sec:partial-body}
We present results on partial human body completion in Fig.~\ref{fig:cond-halves-gen-fig}. Our model can recognize and synthesize partial human bodies in addition to full bodies.

\section{Cloth swapping}
\label{sec:cloth-swap}
In addition to partial bodies, our model can also be used for interactive editing such as swapping clothes as demonstrated in Fig.~\ref{fig:cond-shirts-gen-fig}.

\section{Diversity in predicted poses}
\label{sec:diversity}

Different initial noise maps produce different insertions of the reference person into the scene. We present such diverse insertions predicted by our model for the same input scene and reference person in Fig.~\ref{fig:cond-diverse-gen-fig}.

\section{Failure cases}
\label{sec:failure}

We present common failure modes of our model in Fig.~\ref{fig:cond-failure-gen-fig}.

\begin{itemize}[noitemsep,topsep=0pt,parsep=0pt,partopsep=0pt,leftmargin=2em]
\setlength\itemsep{-0.1em}
\item \textbf{Bad faces and limbs}: Our model often outputs poor face and limb structures (first and second row). This is a result of the first-stage VAE being unable to encode face and limb structures well. This is also a known issue in Stable Diffusion and an active area of development. Training pixel-based diffusion models or improving the auto-encoding quality of the first-stage VAE for humans might alleviate this issue to some degree. These improvements would directly translate to our model.
\item \textbf{Lighting failure}: The harmonization of lightning of the reference person when inserted in the scene fails at times (first row), if the difference is quite large.
\item \textbf{Extreme poses}: Extreme input poses are sometimes not reposed (second row) and the model tries to retain the input pose.
\item \textbf{Blurry samples}: Our model outputs blurry samples at times (third row). Since our training data is primarily videos, we speculate this is due to the motion blur present in the video dataset.
\item \textbf{Object failure}: If the reference person is interacting with a visible object in the input, the model also attempts to insert the object into the scene. This leads to artifacts (fourth row).

\end{itemize}

\section{Societal impact}
\label{sec:societal}

We present a method for affordance-aware human insertion into scenes. We can also hallucinate humans given scene context and vice-versa. The presented research has implications for future work in computer vision, graphics, and robotics. However, our model can be misused to generate malicious content. Similar to Stable Diffusion, the samples generated by our model can be watermarked. Additionally, there's a line of research on detecting fake samples from generative models~\cite{wang2019cnngenerated}, which we encourage the use of. Since our model is trained on internet videos, it inherits several demographic biases present in the data. We believe the research contributions of this work outweigh the negative impacts. 

{\small
\bibliography{egbib}
\bibliographystyle{ieee_fullname}
}